%% file: main.tex
\documentclass[runningheads]{llncs-sup}

\usepackage[table,dvipsnames]{xcolor}

\usepackage{comment}
\usepackage[width=122mm,left=12mm,paperwidth=146mm,height=193mm,top=12mm,paperheight=217mm]{geometry}

\usepackage[normalem]{ulem}
\usepackage{microtype}
\usepackage{wrapfig}
\usepackage{appendix}
\usepackage{epsfig}
\usepackage{graphicx}
\usepackage{amsmath}
\usepackage{amssymb}
\usepackage{pifont}
\usepackage{color}
\usepackage{booktabs}
\usepackage{multirow}
\usepackage{subfigure}
\usepackage{etoolbox}
\usepackage{subfiles}
\usepackage{titletoc}
\usepackage{floatrow}
\usepackage{xurl}
\newfloatcommand{capbtabbox}{table}[][\FBwidth]
\newcommand{\dobib}{ %
    \bibliographystyle{ieee_fullname}
    \bibliography{egbib}
}

\usepackage[pagebackref=true,breaklinks=true,letterpaper=true,colorlinks,bookmarks=false]{hyperref}
\usepackage{cleveref}  %

\newlength{\oldintextsep}
\input{macros.tex}
\begin{document}

\pagestyle{headings}
\mainmatter
\def\ECCVSubNumber{3604}  %

\title{\dataset: A Large-Scale Benchmark for \\Tracking Any Object}

\titlerunning{\dataset: A Large-Scale Benchmark for Tracking Any Object}
\authorrunning{A. Dave, T. Khurana, P. Tokmakov, C. Schmid, D. Ramanan}
\author{Achal Dave\inst{1}\and Tarasha Khurana\inst{1}\and Pavel Tokmakov\inst{1}\\ Cordelia Schmid\inst{2}\and Deva Ramanan\inst{1,3}}
\institute{Carnegie Mellon University \and Inria \qquad\andsameline Argo AI}

\maketitle

\begin{abstract}
For many years, multi-object tracking benchmarks have focused on a handful of categories. Motivated primarily by surveillance and self-driving applications, these datasets provide tracks for people, vehicles, and animals, ignoring the vast majority of objects in the world. By contrast, in the related field of object detection, the introduction of large-scale, diverse datasets (e.g., COCO) have fostered significant progress in developing highly robust solutions. To bridge this gap, we introduce a similarly diverse dataset for Tracking Any Object (\dataset)\footnote{\url{http://taodataset.org/}}. It consists of \TaoTotalVids{} high resolution videos, captured in diverse environments, which are half a minute long on average. Importantly, we adopt a bottom-up approach for discovering a large vocabulary of \TaoAllCategories{} categories, an order of magnitude more than prior tracking benchmarks. To this end, we ask annotators to label objects that move at any point in the video, and give names to them post factum. Our vocabulary is both significantly larger and qualitatively different from existing tracking datasets. To ensure scalability of annotation, we employ a federated approach that focuses manual effort on labeling tracks for those relevant objects in a video (e.g., those that move). We perform an extensive evaluation of state-of-the-art trackers and make a number of important discoveries regarding large-vocabulary tracking in an open-world. In particular, we show that existing single- and multi-object trackers struggle when applied to this scenario in the wild, and that detection-based, multi-object trackers are in fact competitive with user-initialized ones. We hope that our dataset and analysis will boost further progress in the tracking community.
\keywords{datasets, video object detection, tracking}
\end{abstract}

\section{Introduction}

\begin{figure}[t]
  \centering
  \subfigure{
    \includegraphics[height=12.5em]{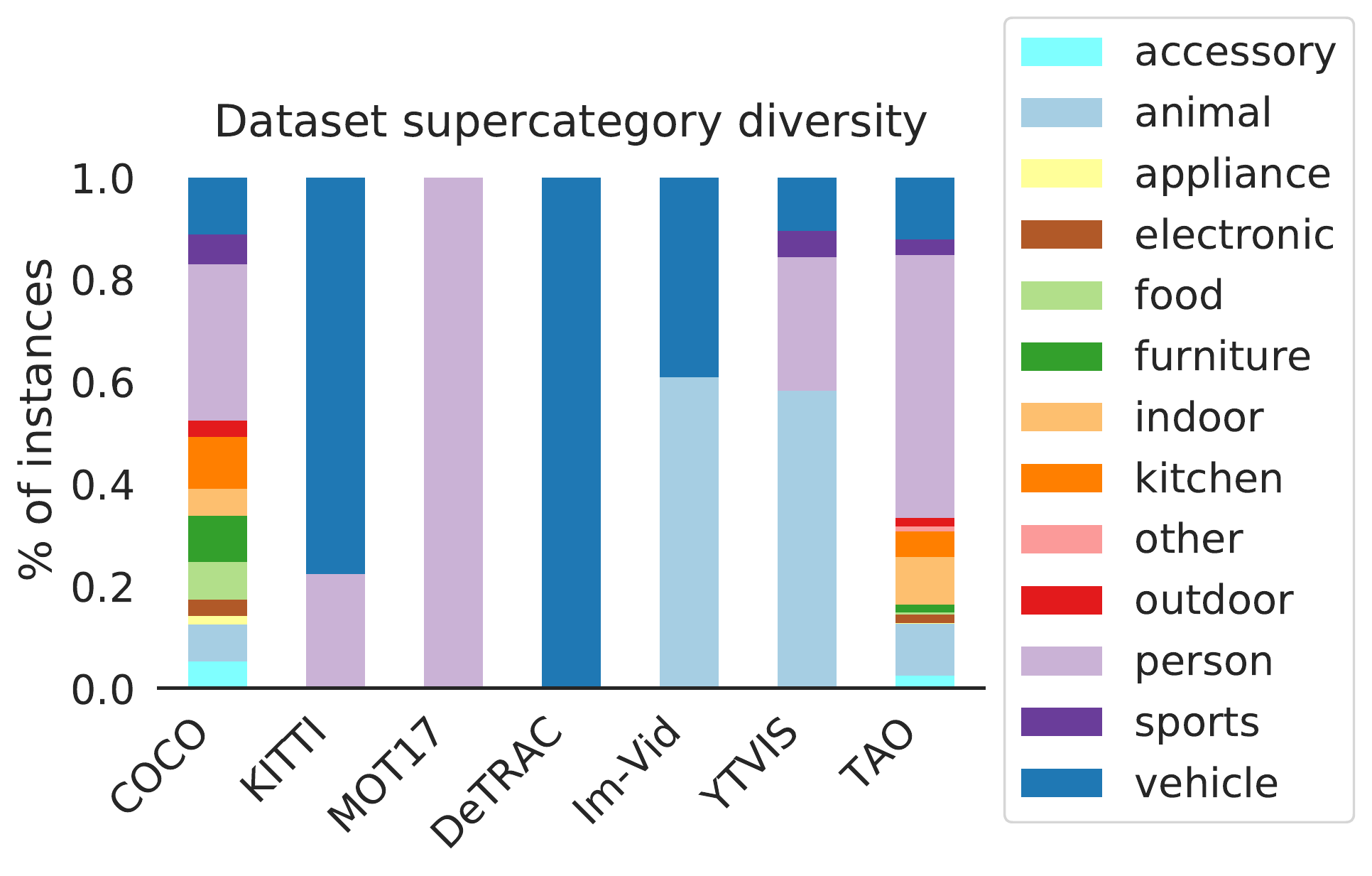}

}\hspace{0.02\linewidth}
\subfigure{
    \raisebox{0.7em}{\includegraphics[height=11.5em]{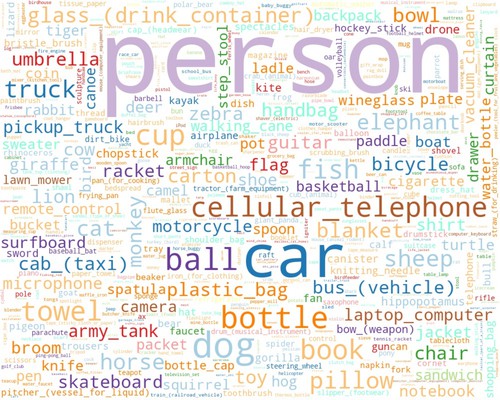}}
    }
    \caption{(left) Super-category distribution in existing multi-object tracking datasets compared to \dataset{} and COCO~\cite{lin2014microsoft}. Previous work focused on people, vehicles and animals. By contrast, our bottom-up category discovery results in a more diverse distribution, covering many small, hand-held objects that are especially challenging from the tracking perspective. (right) Wordcloud of \dataset{} categories, weighted by number of instances, and colored according to their supercategory.}
    \label{fig:cats}
    \vspace{-1em}
\end{figure}

A key component in the success of modern object detection methods was the introduction of large-scale, diverse benchmarks, such as MS COCO~\cite{lin2014microsoft} and LVIS~\cite{gupta2019lvis}. By contrast, multi-object tracking datasets tend to be small~\cite{milan2016mot16,voigtlaender2019mots}, biased towards short videos~\cite{yang2019video}, and, most importantly, focused on a very small vocabulary of categories~\cite{milan2016mot16,voigtlaender2019mots,wen2015ua} (see Table~\ref{tab:dataset_stats}). As can be seen from Figure~\ref{fig:cats}, they predominantly target people and vehicles. Due to the lack of proper benchmarks, the community has shifted towards solutions tailored to the few videos used for evaluation. %
Indeed, Bergmann et al.~\cite{bergmann2019tracking} have recently and convincingly demonstrated that simple baselines perform on par with state-of-the-art (SOTA) multi-object trackers.

\begin{figure*}
    \centering
    \includegraphics[width=\linewidth]{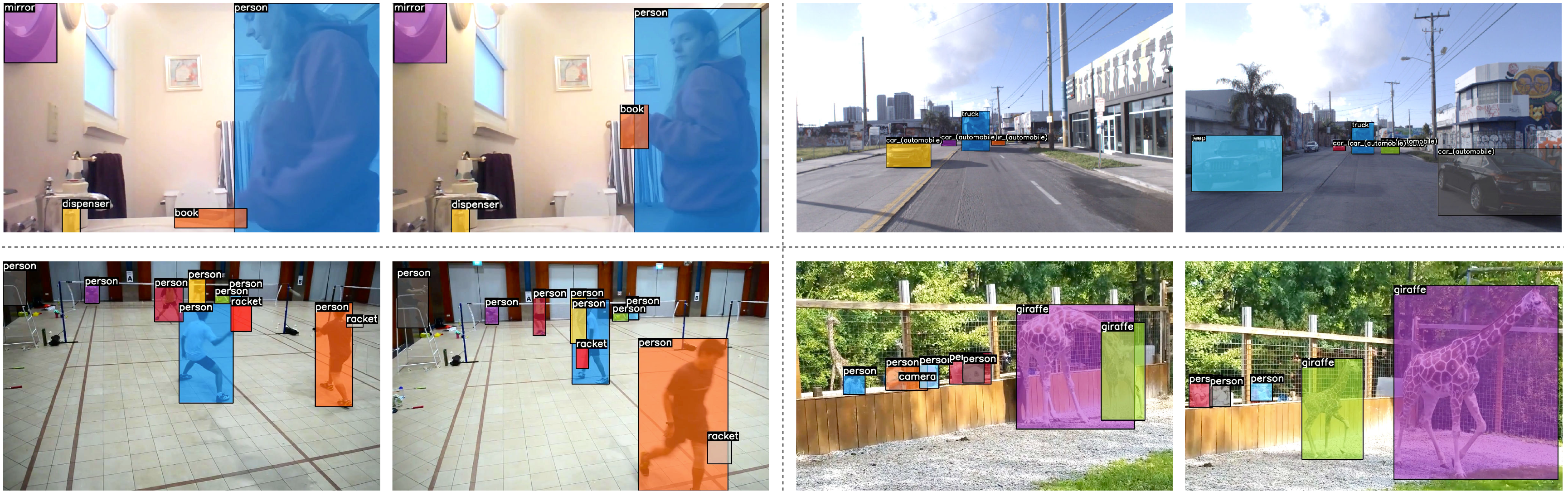}
    \caption{Representative frames from \dataset{}, showing videos sourced from multiple domains with annotations at two different timesteps.}
    \label{fig:tao}
\end{figure*}

In this work we introduce a large-scale benchmark for Tracking Any Object (\dataset). Our dataset features \TaoTotalVids{} high resolution videos captured in diverse environments, which are 30 seconds long on average, and has tracks labeled for \TaoAllCategories{} object categories. We compare the statistics of \dataset{} to existing multi-object tracking benchmarks in Table~\ref{tab:dataset_stats} and Figure~\ref{fig:cats}, and demonstrate that it improves upon them both in terms of complexity and in terms of diversity (see Figure~\ref{fig:tao} for representative frames from \dataset{}). Collecting such a dataset presents three main challenges: (1) how to select a large number of diverse, long, high-quality videos; (2) how to define a set of categories covering all the objects that might be of interest for tracking; and (3) how to label tracks for these categories at a realistic cost. Below we summarize our approach for addressing these challenges. A detailed description of dataset collection is provided in Section~\ref{sec:dataset_collection}.

Existing datasets tend to focus on one or just a few domains when selecting the videos, such as outdoor scenes in MOT~\cite{milan2016mot16}, or road scenes in KITTI~\cite{Geiger2012CVPR}. This results in methods that fail when applied in the wild. To avoid this bias, we construct \dataset{} with videos from as many environments as possible. We include indoor videos from Charades~\cite{sigurdsson2016hollywood}, movie scenes from AVA~\cite{gu2018ava}, outdoor videos from LaSOT~\cite{fan2019lasot}, road-scenes from ArgoVerse~\cite{chang2019argoverse}, and a diverse sample of videos from HACS~\cite{zhao2019hacs} and YFCC100M~\cite{thomee2015yfcc100m}. We ensure all videos are of high quality, with the smallest dimension larger or equal to 480px, and contain at least 2 moving objects.
\Cref{tab:dataset_stats} reports the full statistics of the collected videos, showing that \dataset{} provides an evaluation suite that is significantly larger, longer, and more diverse than prior work.
Note that \dataset{} contains fewer training videos than recent tracking datasets, as we intentionally dedicate the majority of videos for in-the-wild {\em benchmark} evaluation, the focus of our effort. %

\begin{table*}[t]
  \centering
  \caption{Statistics of major multi-object tracking datasets.
   \dataset{} is by far the largest dataset in terms of the number of categories, and the total duration of videos used for evaluation. In addition, we ensure that each video is challenging (long, containing several moving objects) and of high quality.
   }
   \resizebox{\linewidth}{!}{
  \begin{tabular}{lcccccccc}
    \toprule
    Dataset & Classes &
    \multicolumn{2}{c}{
      \begin{tabular}{cc}
        \multicolumn{2}{c}{Videos} \\
        Eval. & Train
      \end{tabular}
    } &
    \begin{tabular}{c}Avg\\length (s)\end{tabular} &
    \begin{tabular}{c}Tracks \\ / video\end{tabular} &
    \begin{tabular}{c}Min \\resolution\end{tabular} &
    \begin{tabular}{c}Ann.\\fps\end{tabular} &
    \begin{tabular}{c}Total Eval\\length (s)\end{tabular} \\\midrule
    MOT17~\cite{milan2016mot16}      & 1    & 7 & 7     & 35.4 & 112  & 640x480 & 30 & 248 \\
    KITTI~\cite{Geiger2012CVPR}    & 2    & 29 & 21    & 12.6 & 52 & 1242x375 & 10 & 365 \\
    UA-DETRAC~\cite{wen2015ua}       & 4    & 40 & 60  & 56 & 57.6 & 960x540  & 5   & 2,240 \\
    \imvid{} ~\cite{russakovsky2015imagenet} & 30 & 1,314 & 4,000 & 10.6 & 2.4 & 480x270 & {\raise.17ex\hbox{$\scriptstyle\sim$}}25 & 13,928 \\
    YTVIS ~\cite{yang2019video}  & 40   & 645 & 2,238  & 4.6   & 1.7 & 320x240 & 5 & 2,967 \\
    \dataset{} (Ours) & \TaoAllCategories & \TaoEvaluationVids{} & \TaoTrainVids{} & \TaoAvgVidLength  & \TaoAvgNumTracks & 640x480 & 1 & \TaoEvalTotalVidLength \\
    \bottomrule
  \end{tabular}
   }
  \label{tab:dataset_stats}
\end{table*}

Given the selected videos, we must choose {\em what} to annotate. Most datasets are constructed with a top-down approach, where categories of interest are pre-defined by benchmark curators. That is, curators first select the subset of categories deemed relevant for the task, and then collect images or videos expressly for these categories~\cite{deng2009imagenet,lin2014microsoft,valmadre2018long}. This approach naturally introduces curator bias. An alternative strategy is bottom-up, open-world {\em discovery} of what objects are present in the data. Here, the vocabulary emerges post factum~\cite{gu2018ava,gupta2019lvis,zhou2017scene}, an approach that dates back to LabelMe~\cite{russell2008labelme}. %
Inspired by this line of work, we devise the following strategy to discover an ontology of objects relevant for tracking: first annotators are asked to label {\em all} objects that either move by themselves or are moved by people. They then give names to the labeled objects, resulting in a vocabulary that is not only significantly larger, but is also qualitatively different from that of any existing tracking dataset (see Figure~\ref{fig:cats}). To facilitate training of object detectors, that can be later used by multi-object trackers on our dataset, we encourage annotators to choose categories that exists in the LVIS dataset~\cite{gupta2019lvis}. If no appropriate category can be found in the LVIS vocabulary, annotators can provide free-form names (see Section~\ref{sec:annotation_pipeline} for details).

Exhaustively labeling tracks for such a large collection of objects in \TaoTotalVids{} long videos is prohibitively expensive. Instead, we extend the federated annotation approach proposed in~\cite{gupta2019lvis} to the tracking domain. In particular, we ask the annotators to label tracks for up to 10 objects in every video. We then separately collect exhaustive labels for every category for a subset of videos, indicating whether all the instances of the category have been labeled in the video. During evaluation of a particular category, we use only videos with exhaustive labels for computing precision and all videos for computing recall.
This allows us to reliably measure methods' performance at a fraction of the cost of exhaustively annotating the videos.
We use the LVIS federated mAP metric~\cite{gupta2019lvis} for evaluation, replacing 2D IoU with 3D IoU~\cite{yang2019video}.
For detailed comparisons, we further report the standard MOT challenge~\cite{milan2016mot16} metrics in \Cref{sec:appendix_mota}.

Equipped with \dataset{}, we set out to answer several questions about the state of the tracking community. In particular, in Section~\ref{sec:analysis} we report the following discoveries:
(1)~SOTA trackers struggle to generalize to a large vocabulary of objects, particularly for infrequent object categories in the tail;
(2)~while trackers work significantly better for the most-explored category of people, tracking people in diverse scenarios (e.g., frequent occlusions or camera motion) remains challenging;
(3)~when scaled to a large object vocabulary, multi-object trackers become competitive with user-initialized trackers, despite the latter being provided with a ground truth initializations.
 We hope that these insights will help to define the most promising directions for future research.

\section{Related work}
\label{sec:related}
The domain of object tracking is subdivided based on the way the tracks are initialized. Our work falls into the multi-object tracking category, where all the objects out of a fixed vocabulary of classes have to be detected and tracked. Other formulations include user-initialized tracking, and saliency-based tracking. In the remainder of this section we will first review the most relevant benchmarks datasets in each of these areas, and then discuss SOTA methods for multi-object and user-initialized tracking.

\subsection{Benchmarks}
\label{sec:related_benchmarks}
\smallsec{Multi-object tracking (MOT)} is the task of tracking an unknown number of objects from a known set of categories. Most MOT benchmarks~\cite{fisher2001context,Geiger2012CVPR,milan2016mot16,wen2015ua} focus on either people or vehicles (see Figure~\ref{fig:cats}), motivated by surveillance and self-driving applications. Moreover, they tend to include only a few dozen videos, captured in outdoor or road environments, encouraging methods that are overly adapted to the benchmark and do not generalize to different scenarios (see Table~\ref{tab:dataset_stats}%
). In contrast, \dataset{} focuses on diversity both in the category and visual domain distribution, resulting in a realistic benchmark for tracking {\em any} object.

Several works have attempted to extend the MOT task to a wider vocabulary of categories. In particular, the \imvid{}~\cite{russakovsky2015imagenet} benchmark provides exhaustive trajectories annotations for objects of 30 categories in 1314 videos. While this dataset is both larger and more diverse that standard MOT benchmarks, videos tend to be relatively short and the categories cover only animals and vehicles. The recent YTVIS dataset~\cite{yang2019video} has the most broad vocabulary to date, covering 40 classes, but the majority of the categories still correspond to people, vehicles and animals. Moreover, the videos are 5 seconds long on average, making the tracking problem considerably easier in many cases. Unlike previous work, we take a bottom-up approach for defining the vocabulary. This results in not only the largest set of categories among MOT datasets to date, but also in a qualitatively different category distribution. In addition, our dataset is over 7 times larger than YTVIS in the number of frames.  %
The recent VidOR dataset~\cite{shang2019annotating} explores Video Object Relations, including tracks for a large vocabulary of objects. But, since ViDOR focuses on relations rather than tracks, object trajectories tend to be missing or incomplete, making it hard to repurpose for tracker benchmarking.
In contrast, we ensure TAO maintains high quality for both accuracy and completeness of labels (see Appendix~\ref{sec:appendix-quality} for a quantitative analysis).

Finally, several recent works have proposed to label masks instead of bounding boxes for benchmarking multi-object tracking~\cite{voigtlaender2019mots,yang2019video}. In collecting \dataset{} we made a conscious choice to prioritize scale and diversity of the benchmark over pixel-accurate labeling. Instance mask annotations are significantly more expensive to collect than bounding boxes, and we show empirically that tracking at the box level is already a challenging task that current methods fail to solve.

\smallsec{User-initialized tracking} forgoes a fixed vocabulary of categories altogether and instead relies on the user to provide bounding box annotations for the objects that need to be tracked at test time~\cite{fan2019lasot,huang2018got,kristan2016novel,valmadre2018long,wu2013online}. The benchmarks in this category tend to be larger and more diverse than their MOT counterparts, but most of them still offer a tradeoff between the number of videos in the benchmarks and the average length of the videos (see \Cref{sec:appendix_tao_stats}). Moreover, even if the task itself is category-agnostic, empirical distribution of categories in the benchmarks tends to be heavily skewed towards a few common objects. We study whether this bias in category selection results in methods failing to generalize to more challenging objects by evaluating state-of-the-art user-initialized trackers on \dataset{} in Section~\ref{sec:results_user_init}.

\smallsec{Semi-supervised video object segmentation} differs from user-initialized tracking in that both the input to the tracker and the output are object masks, not boxes~\cite{perazzi2016benchmark,xu2018youtube}. As a result, such datasets are a lot more expensive to collect, and videos tend to be extremely short. The main focus of the works in this domain~\cite{caelles2017one,khoreva2019lucid,voigtlaender2017online} is on accurate mask propagation, not solving challenging identity association problems, thus their effort is complementary to ours.

\smallsec{Saliency-based tracking} is an intriguing direction towards open-world tracking, where the objects of interest are defined not with a fixed vocabulary of categories, or manual annotations, but with bottom-up, motion-~\cite{ochs2013segmentation,perazzi2016benchmark} or appearance-based~\cite{caelles20192019,wang2019learning} saliency cues.
Our work similarly uses motion-based saliency to define a comprehensive vocabulary of categories, but presents a significantly larger benchmark with
class labels for each object, enabling the use and evaluation of large-vocabulary object recognition approaches.

\subsection{Algorithms}
\label{sec:related_algorithms}
\smallsec{Multi-object trackers} can be categorized into people and multi-category trackers. The former have been mainly developed on the MOT benchmark~\cite{milan2016mot16} and follow the tracking-by-detection paradigm, linking outputs of person detectors in an offline, graph-based framework~\cite{berclaz2006robust,berclaz2011multiple,breitenstein2009robust,ess2008mobile}. These methods mainly differ in the way they define the edge cost in the graph. Classical approaches use overlap between detections in consecutive frames~\cite{jiang2007linear,pirsiavash2011globally,zhang2008global}. More recent methods define edge costs based on appearance similarity~\cite{milan2017online,ristani2018features}, or motion-based models~\cite{alahi2016social,chen2018real,choi2010multiple,leal2014learning,ren2018collaborative,scovanner2009learning}. Very recently, Bergmann et al.~\cite{bergmann2019tracking} proposed a simple baseline approach that performs on par with SOTA people trackers, which repurposes an object detector's bounding box regression capability to predict the position of an object in the next frame. Notice that all these methods have been developed and evaluated on the relatively small MOT dataset, which consists of 14 videos captured in very similar environments.
By contrast, \dataset{} provides a much richer, more diverse set of videos, encouraging trackers more robust to tracking challenges such as occlusion and camera motion.

The more general multi-object tracking scenario is usually studied using \imvid{}~\cite{russakovsky2015imagenet}. Methods in this group also use offline, graph-based optimization to link frame-level detections into tracks. To define the edge potentials, in addition to bounding box overlap, Feichtenhofer et al.~\cite{feichtenhofer2017detect} propose to use a similarity embedding, which is learned jointly with the detector. Alternatively, Kang et al.~\cite{kang2017object} directly predict short tubelets, and Xiao et al.~\cite{xiao2018video} incorporate a spatio-temporal memory module inside a detector. Inspired by~\cite{bergmann2019tracking}, we show that a simple baseline approach, relying on the Viterbi algorithm for linking detections across frames~\cite{feichtenhofer2017detect,gkioxari2015finding}, performs on par with the methods mentioned above on \imvid{}. We then use this baseline for evaluating generic multi-object tracking on \dataset{} in Section~\ref{sec:results_mot}, and demonstrate that it struggles when faced with a large vocabulary and a diverse data distribution.

\smallsec{User-initialized trackers} tend to rely on a Siamese network architecture that was first introduced for signature verification~\cite{bromley1994signature}, and later adapted for tracking~\cite{bertinetto2016fully,danelljan2017eco,held2016learning,tao2016siamese}. They learn a patch-level distance embedding and find the closest patch to the one annotated in the first frame in the following frames. To simplify the matching problem, state-of-the-art approaches  limit the search space to the region in which the object was localized in the previous frame. Recently there have been several attempts to introduce some ideas from CNN architectures for object detection into Siamese trackers. In particular, Li et al.~\cite{li2018high} use the similarity map obtained by matching the object template to the test frame as input to an RPN-like module adapted from Faster-RCNN~\cite{ren2015faster}. Later this architecture was extended by introducing hard negative mining and template updating~\cite{zhu2018distractor}, as well as mask prediction~\cite{wang2019fast}. In another line of work, Siamese-based trackers have been augmented with a target discrimination module to improve their robustness to distractors~\cite{bhat2019learning,danelljan2019atom}. We evaluate several state-of-the-art methods in this paradigm for which public implementation is available~\cite{bhat2019learning,danelljan2019atom,danelljan2017eco,li2019siamrpn++,wang2019fast} on \dataset{}, and demonstrate that they achieve only a moderate improvement over our multi-object tracking baseline, despite being provided with a ground truth initialization for each track (see Section~\ref{sec:results_user_init} for details).

\section{Dataset design}
\label{sec:dataset_design}

Our primary goal in this work is collecting a large-scale dataset of videos with a diverse vocabulary of labeled object tracks for evaluating trackers in the wild.
This requires designing a strategy for (1) video collection, (2) vocabulary discovery, (3) scalable annotation, and (4) evaluation.
We detail our strategies for (2-4) in this section, and defer the discussion of video collection to Section~\ref{sec:video_selection}.

\smallsec{Category discovery.}
Rather than manually defining a set of categories, we discover an object vocabulary from unlabeled videos which span diverse operating domains.
Our goal is to focus on \textit{dynamic} objects in the world. Towards this end, we ask annotators to mark all objects that \textit{move} in our collection of videos, without any object vocabulary in mind.
We then construct a vocabulary by giving names for all the discovered objects, following the recent trend for open-world dataset collection~\cite{gupta2019lvis,zhou2017scene}. In particular, annotators are asked to provide a free-form name for every object, but are encouraged to select a category from the LVIS~\cite{gupta2019lvis} vocabulary whenever possible. We detail this process further in Section~\ref{sec:annotation_categorization}.

\smallsec{Federation.}
Given this vocabulary, one option might be exhaustively labelling all instances of each category in all videos. Unfortunately, exhaustive annotation of a large vocabulary is expensive, even for images, as noted in~\cite{gupta2019lvis}.
We choose to use our labeling budget instead on collecting a large-scale, diverse dataset, by
extending the federated annotation protocol of~\cite{gupta2019lvis} from image datasets to videos.
Rather than labeling every video $v$ with every category $c$, we define three subsets of our dataset for each category: $P_c$, which contains videos where all instances of $c$ are labeled, $N_c$, videos with no instance of $c$ present in the video, and $U_c$, videos where \textit{some} instances of $c$ are annotated.
Videos not belonging to any of these subsets are ignored when evaluating category $c$.
For each category $c$, we only use videos in $P_c$ and $N_c$ to measure the \textit{precision} of trackers, and videos in $P_c$ and $U_c$ to measure recall.
We describe how to define $P_c$, $N_c$, and $U_c$ in Section~\ref{sec:annotation_federation}.

\smallsec{Granularity of annotations.}
To collect \dataset{}, we choose to prioritize scale and diversity of the data at the cost of annotation granularity. In particular, we label tracks at 1 frame per second with bounding box labels but don't annotated segmentation masks. This allows us to label \TaoAllCategories{} categories in \TaoTotalVids{} videos at a relatively modest cost. Our decision is motivated by the observation of~\cite{valmadre2018long} that dense frame labeling does not change the relative performance of the methods.

\smallsec{Evaluation and metric.}
Traditionally, multi-object tracking datasets use either the CLEAR MOT metrics~\cite{bernardin2008evaluating,Geiger2012CVPR,milan2016mot16} or a 3D intersection-over-union (IoU) based metric~\cite{russakovsky2015imagenet,yang2019video}. We report the former in \Cref{sec:appendix_mota} (introducing modifications for large-vocabularies of classes, including multi-class aggregation and federation), but focus our experiments on the latter.
To formally define 3D IoU, let $G = \{g_1, \dots, g_T\}$ and $D = \{d_1, \dots, d_T\}$  be a groundtruth and predicted track for a video with $T$ frames. 3D IoU is defined as:
$\text{IoU}_{\text{3d}}(D, G) = \frac{\sum_{t=1}^T g_t \cap d_t}{\sum_{t=1}^T g_t \cup d_t}.$
If an object is not present at time $t$, we assign $g_t$ to an empty bounding box, and similarly for a missing detection.
We choose 3D IoU (with a threshold of 0.5) as the default metric for \dataset{}, and provide further analysis in \Cref{sec:appendix-metrics}.

Similar to standard object detection metrics, (3D) IoU together with (track) confidence can be used to compute mean average precision across categories. For methods that provide a score for each frame in a track, we use the average frame score as the track score. %
Following~\cite{gupta2019lvis}, we measure precision for a category $c$ in video $v$ only if all instances of the category are verified to be labeled in it.  %
\vspace{-1em}

\section{Dataset collection}
\label{sec:dataset_collection}

\subsection{Video selection}
\label{sec:video_selection}
Most video datasets focus on one or a few operating domains. For instance, MOT benchmarks \cite{milan2016mot16} correspond to urban, outdoor scenes featuring crowds of people, whereas AVA~\cite{gu2018ava} is sourced from produced films, typically capturing actors with close shots in carefully staged scenes. %
As a result, methods developed on any single dataset (and hence domain) fail to generalize in the wild. To avoid this bias, we constructed \dataset{} by selecting videos from a variety of existing video benchmarks to ensure diversity of scenes and objects.

\smallsec{Diversity.} In particular, we used datasets for action recognition, self-driving cars,  user-initialized tracking, as well as in-the-wild Flickr videos. In the action recognition domain we selected 3 datasets: Charades~\cite{sigurdsson2016hollywood}, AVA~\cite{gu2018ava}, and HACS~\cite{zhao2019hacs}. Charades features complex human-human and human-object interactions, but all videos are indoor with limited camera motion. In contrast, AVA has a much wider variety of scenes and cinematographic styles but is scripted. HACS provides unscripted, in-the-wild videos. These action datasets are naturally focused on people and objects with which people interact. To include other animals and vehicles, we also source clips from LaSOT~\cite{fan2019lasot} (a benchmark for user-initialized tracking), BDD~\cite{bdd100k} and ArgoVerse~\cite{chang2019argoverse} (benchmarks for self-driving cars). LaSOT is a diverse collection whereas BDD and ArgoVerse consist entirely of outdoor, urban scenes. Finally we sample %
in-the-wild videos from the YFCC100M~\cite{thomee2015yfcc100m} Flickr collection.

\smallsec{Quality.} The videos are automatically filtered to remove short videos and videos with a resolution below 480p. For longer videos, as in AVA, we use~\cite{LokocKSMC19Viret} to extract scenes without shot changes. In addition, we manually reviewed each sampled video to ensure it is high quality: i.e., we removed grainy videos as well as videos with excessive camera motion or shot changes. Finally, to focus on the most challenging tracking scenarios, we only kept videos that contain at least 2 moving objects. The full statistics of the collected videos are provided in Table~\ref{tab:dataset_stats}. We point out that many prior video datasets tend to limit one or more quality dimensions (in terms of resolution, length, or number of videos) in order to keep evaluation and processing times manageable. In contrast, we believe that in order to truly enable tracking in the open-world, we need to appropriately scale benchmarks. %

\begin{figure*}[t]
    \centering
    \includegraphics[width=\linewidth]{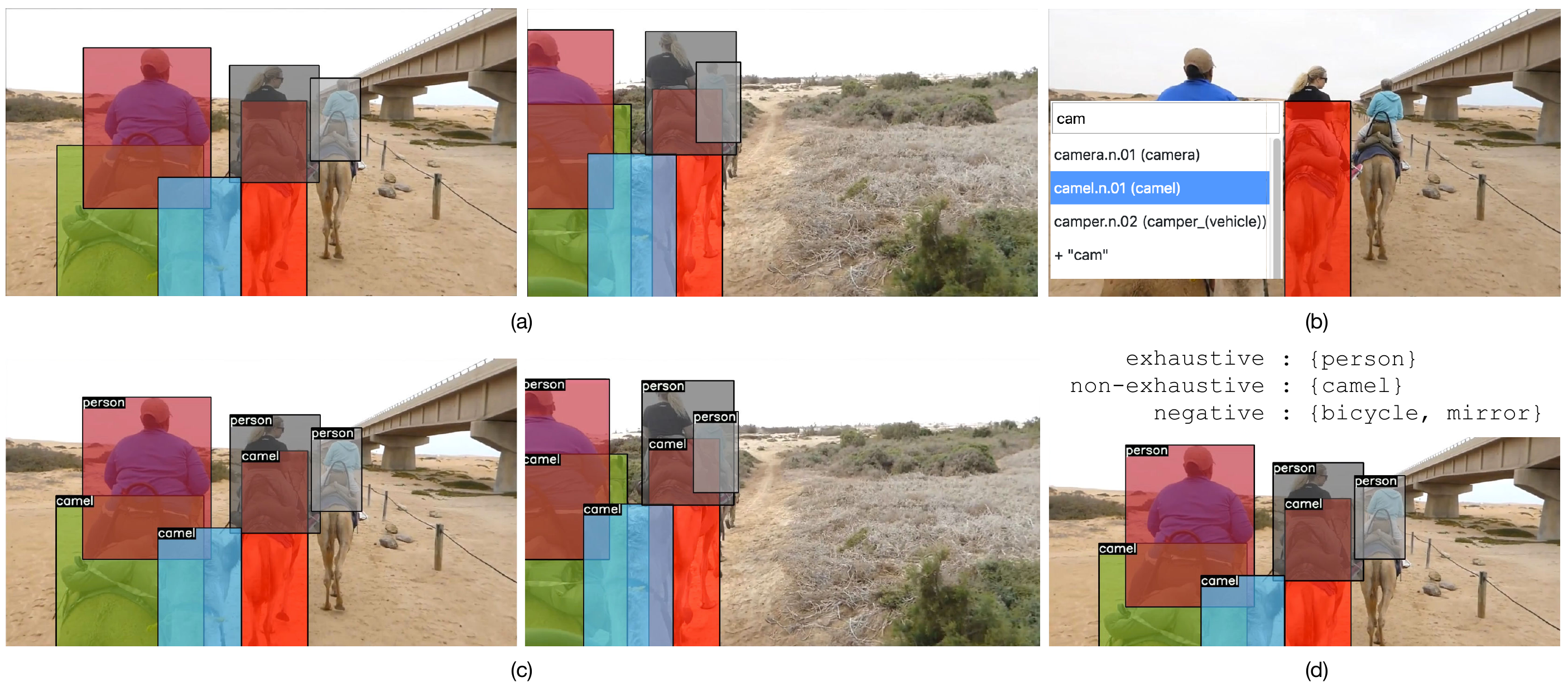}
    \caption{Our federated video annotation pipeline. First (a), annotators mine and track moving objects. Second (b), annotators categorize tracks using categories from the LVIS vocabulary or free-form text, producing the labeled tracks (c). Finally, annotators identify categories that are exhaustively annotated or verified to be absent. In this example (d), `person's are identified as being exhaustively annotated, `camel's are present but not exhaustively annotated and `bicycle's and `mirror's are verified as absent. Such federated labels allow one to accurately penalize false-positives and missed detections for exhaustively annotated and verified categories.
    }
    \label{fig:pipeline}
\end{figure*}

\subsection{Annotation pipeline}
\label{sec:annotation_pipeline}
Our annotation pipeline is illustrated in Figure~\ref{fig:pipeline}. We designed it to separate low-level tracking  from high-level semantic labeling. As pointed out by others~\cite{barriuso2012notes}, semantic labeling can be subtle and error-prone because of ambiguities and corner-cases that arise in category boundaries. By separating tasks into low vs high-level, we are able to take advantage of unskilled annotators for the former and highly-vetted workers for the latter.

\subsubsection{Object mining and tracking.}
\label{sec:annotation_mining}
We combine  object mining and track labeling into a single stage of annotation. Given the set of videos described above, we ask annotators to mark {\em objects that  move at any point in the video}. To avoid overspending our annotation budget on a few crowded videos, we limited the number of labeled object per video to 10. %
Note that this stage is \textit{category-agnostic}: annotators are not instructed to look for objects from any specific vocabulary, but instead use motion as a {\em saliency} cue for mining relevant objects. They are then asked to track these objects throughout the video, and label them with bounding boxes at 1 frame-per-second. Finally, the tracks are verified by one independent annotator. This process is illustrated in Figure~\ref{fig:pipeline}, where we can see that 6 objects are discovered and tracked.

\subsubsection{Object categorization.}
\label{sec:annotation_categorization}
Next, we collected category labels for objects discovered in the previous stage and simultaneously constructed the dataset vocabulary. We focus on the large vocabulary from the LVIS~\cite{gupta2019lvis} object detection dataset, which contains 1,230 synsets discovered in a bottom-up manner similar to ours. Doing so also allows us to make use of LVIS as a training set of relevant object detectors (which we later use within a tracking pipeline to produce strong baselines - \Cref{sec:analysis_methods}). Because maintaining a mental list of 1,230 categories is challenging even for expert annotators, we use an auto-complete annotation interface to suggests categories from the LVIS vocabulary (Figure~\ref{fig:pipeline} (b)).
The autocomplete interface displays classes with a matching synset (e.g., ``person.n.01"), name, synonym, and finally those with a matching definition.
Interestingly, we find that some objects discovered in \dataset{}, such as ``door'' or ``marker cap'', do not exist in LVIS. To accommodate such important exceptions, we allow annotators to label objects with free-form text if they do not fit in the LVIS vocabulary.

Overall, annotators labeled \TaoLvisObjects{} objects (\TaoLvisObjectsPerc{}\%) with \TaoLvisCategories{} LVIS categories, and \TaoNonLvisObjects{} objects (\TaoNonLvisObjectsPerc{}\%) with \TaoNonLvisCategories{} free-form categories. We use the \TaoLvisCategories{} LVIS categories for MOT experiments (because detectors can be trained on LVIS), but use all categories for user-initialized tracking experiments in \Cref{sec:appendix-all-categories}.

\subsubsection{Federated ``exhaustive'' labeling.}
\label{sec:annotation_federation}
Finally, we ask annotators to verify which categories are exhaustively labeled for each video.
Specifically, for each category $c$ labeled in video $v$, we ask annotators whether all instances of $c$ are labeled.
In \Cref{fig:pipeline}, after this stage, annotators marked that `person' is exhaustively labeled, while `camel' is not.
Next, we show annotators a sampled subset of categories that are not labeled in the video, and ask them to indicate categories which are absent in the video.
In \Cref{fig:pipeline}, annotators indicated that `bicycle' and `mirror' are absent.

\subsection{Dataset splits}
\label{sec:dataset_splits}

We intend for \dataset{} to be used primarily as an \textit{evaluation} benchmark.
We split \dataset{} into three subsets: train, validation and test, containing
\TaoTrainVids{}, \TaoValVids{} and \TaoTestVids{} videos respectively.
Typically, `train' splits tend to be larger than `val' and `test'. We choose to make TAO's training set small for several reasons. Firstly, the primary goal of TAO is to reliably benchmark trackers in-the-wild. Secondly, most MOT systems are modularly trained using image-based detectors with hyper-parameter tuning of the overall tracking system. In our case, we ensure the train set is sufficiently large for hyper-parameter tuning, and ensure that our large-vocabulary is aligned with large-vocabulary image datasets (e.g., LVIS). This allows us to devote most of our annotation budget for large-scale `val' and held-out `test' sets."
We ensure that the videos in train, validation and test are well-separated.
As an example, we ensure that each subject in the Charades dataset appears in only one of the train, validation or test sets.
We provide further details on split construction in Appendix~\ref{sec:appendix_splits}.

\section{Analysis of state-of-the-art trackers}
\label{sec:analysis}
We now use \dataset{} to analyze how well existing multi- and single-object trackers perform in the wild and when they fail.
We tune the hyperparameters of each tracking approach on the `train' set, and report results on the `val' set.
To capitalize on existing object detectors, we evaluate using the~\TaoLvisCategories{} LVIS categories in \dataset{}. We begin by shortly describing the methods used in our analysis.

\subsection{Methods}
\label{sec:analysis_methods}
\smallsec{Detection.}
We analyze how well state-of-the-art object detectors perform on our dataset. To this end, we present results using a standard Mask R-CNN~\cite{ren2015faster} detector trained using~\cite{wu2019detectron2} in \Cref{sec:results_detection}.

\setlength{\oldintextsep}{\intextsep}
\setlength\intextsep{1.5em}
\let\oldFBaskip\FBaskip
\renewcommand\FBaskip{-2.2em}
\begin{wraptable}{r}{0.5\textwidth}
  \centering
  \caption{\imvid{} detection and track mAP; see text (left) for details.}
  \resizebox{\textwidth}{!}{
  \begin{tabular}{lc@{\hskip 1em}c@{\hskip 1.5em}c}
    \toprule
               & Viterbi  & Det mAP & Track mAP \\\midrule
    Detection  &           & 73.4\rlap{ ~\cite{xiao2018video}}  & -     \\
    D\&T~\cite{feichtenhofer2017detect}
               & \cmark    & 79.8   & -     \\
    STMN~\cite{xiao2018video}
               & \cmark    & 79.0   & 60.4     \\\midrule
    Detection  & \cmark    & 79.2   & 60.3  \\
    \bottomrule
  \end{tabular}
  }
    \label{tab:imagenetvid}
\end{wraptable}
\setlength{\intextsep}{\oldintextsep}
\renewcommand\FBaskip{\oldFBaskip}

\smallsec{Multi-Object Tracking.}
We analyze SOTA multi-object tracking methods on \imvid{}, the largest vocabulary dataset prior to \dataset{}.
We first clarify whether such approaches improve detection or tracking.
\Cref{tab:imagenetvid} reports the standard \imvid{} Detection mAP and Track mAP.
The `Detection' row corresponds to a detection-only baseline widely reported by prior work~\cite{xiao2018video,feichtenhofer2017detect,Zhu2017FlowGuidedFA}.
D\&T~\cite{feichtenhofer2017detect} and STMN~\cite{xiao2018video} are spatiotemporal architectures that produce SOTA improvements of 6-7\% in detection mAP over a per-frame detector.
However, both D\&T and STMN post-process their per-frame outputs using the Viterbi algorithm, which iteratively links and re-weights the confidences of per-frame detections (see~\cite{gkioxari2015finding} for details).
{\em When the same post-processing is applied to a single-frame detector, one achieves nearly the same performance gain (Table~\ref{tab:imagenetvid}, last row)}.

Our analysis reinforces the bleak view of multi-object tracking progress suggested by~\cite{bergmann2019tracking}:
while ever-more complex approaches have been proposed for the task, their improvements are often attributable to simple, baseline strategies.
To foster meaningful progress on \dataset{}, we evaluate a number of strong baselines in this work.
We evaluate a powerful single-frame detector trained on LVIS~\cite{gupta2019lvis} and COCO~\cite{lin2014microsoft}, followed by two linking methods: SORT~\cite{Bewley2016_sort}, a simple, online linker initially proposed for tracking people, and the Viterbi post-processing step used by~\cite{feichtenhofer2017detect,xiao2018video}, in~\Cref{sec:results_mot}.

\smallsec{Person detection and tracking.}
Detecting and tracking people have been a distinct focus in the multi-object tracking community.
\Cref{sec:results_people} compares the above baselines to a recent SOTA people-tracker~\cite{bergmann2019tracking}.

\smallsec{User-initialized tracking.}
We additionally present results using user-initialized trackers. We evaluate several recent methods for which public implementation is available~\cite{bhat2019learning,danelljan2019atom,danelljan2017eco,li2019siamrpn++,wang2019fast}.
Unfortunately, these trackers do not provide a class label for the objects they are tracking, and cannot directly be compared to multi-object trackers.
However, these trackers \textit{can} be evaluated with an oracle classifier, allowing us to directly compare their accuracy with the methods that simultaneously detect and track objects.

\smallsec{Oracles.}
Finally, to disentangle the complexity of object classification and tracking, we use two oracles. The first, a class oracle, computes the best matching between predicted and groundtruth tracks in each video. Predicted tracks that match to a groundtruth track with 3D IoU $>0.5$ are assigned the category of their matched groundtruth track.
Tracks that do not match to a groundtruth track are not modified, and are treated as false positives.
This allows us to evaluate the performance of trackers assuming the semantic \textit{classification} task is solved.

The second oracle computes the best possible assignment of per-frame detections to tracks, by comparing them with groundtruth. When doing so, class predictions for each detection are held constant. Any detections that are not matched are discarded. This oracle allows us to analyze the best performance we could expect given a fixed set of detections.

\subsection{Results}
\label{sec:results_all}

\smallsec{How hard is object detection on TAO?}
\label{sec:results_detection}
We start by assessing the difficulty of the detection task on TAO. To this end we evaluate the SOTA object detector~\cite{he2017mask} using detection mAP. We train this model on a combination of LVIS and COCO, finding that training on LVIS alone led to a model that struggles to detect people. The final model achieves an mAP of 27.1 on \dataset{} val at IoU 0.5, suggesting that single-frame detection is challenging on \dataset{}.

\subsubsection{Do multi-object trackers generalize to TAO?}
\label{sec:results_mot}
\begin{figure}
  \begin{floatrow}
  \capbtabbox{%
  \begin{tabular}{lcc@{\hskip 1em}c}
    \toprule
           & \multicolumn{2}{c}{Oracle} & \\
    Method                  & Class & Track & Track mAP \\\midrule
    Detection
      &        &        & 0.6 \\\midrule
    Viterbi \cite{feichtenhofer2017detect,gkioxari2015finding}
      &        &        & 6.3  \\
    SORT \cite{Bewley2016_sort}
      &        &        & 13.2 \\
    Detection
      &        & \cmark & 31.5 \\\midrule
    Viterbi  \cite{feichtenhofer2017detect,gkioxari2015finding}
      & \cmark &        & 15.7 \\
    SORT  \cite{Bewley2016_sort}
      & \cmark &        & 30.2 \\
    Detection
      & \cmark & \cmark & 83.6 \\
    \bottomrule
  \end{tabular}
  }{%
    \caption{SORT~\cite{Bewley2016_sort} and Viterbi linking~\cite{feichtenhofer2017detect,gkioxari2015finding} provide strong baselines on \dataset{}, but detection and tracking remain challenging. Relabeling and linking detections from current detectors using the class and track oracles is sufficient to achieve high performance, suggesting a pathway for progress on \dataset{}.
     \label{tab:mot_tao}
    }%
  }
  \ffigbox{%
    \includegraphics[width=\linewidth]{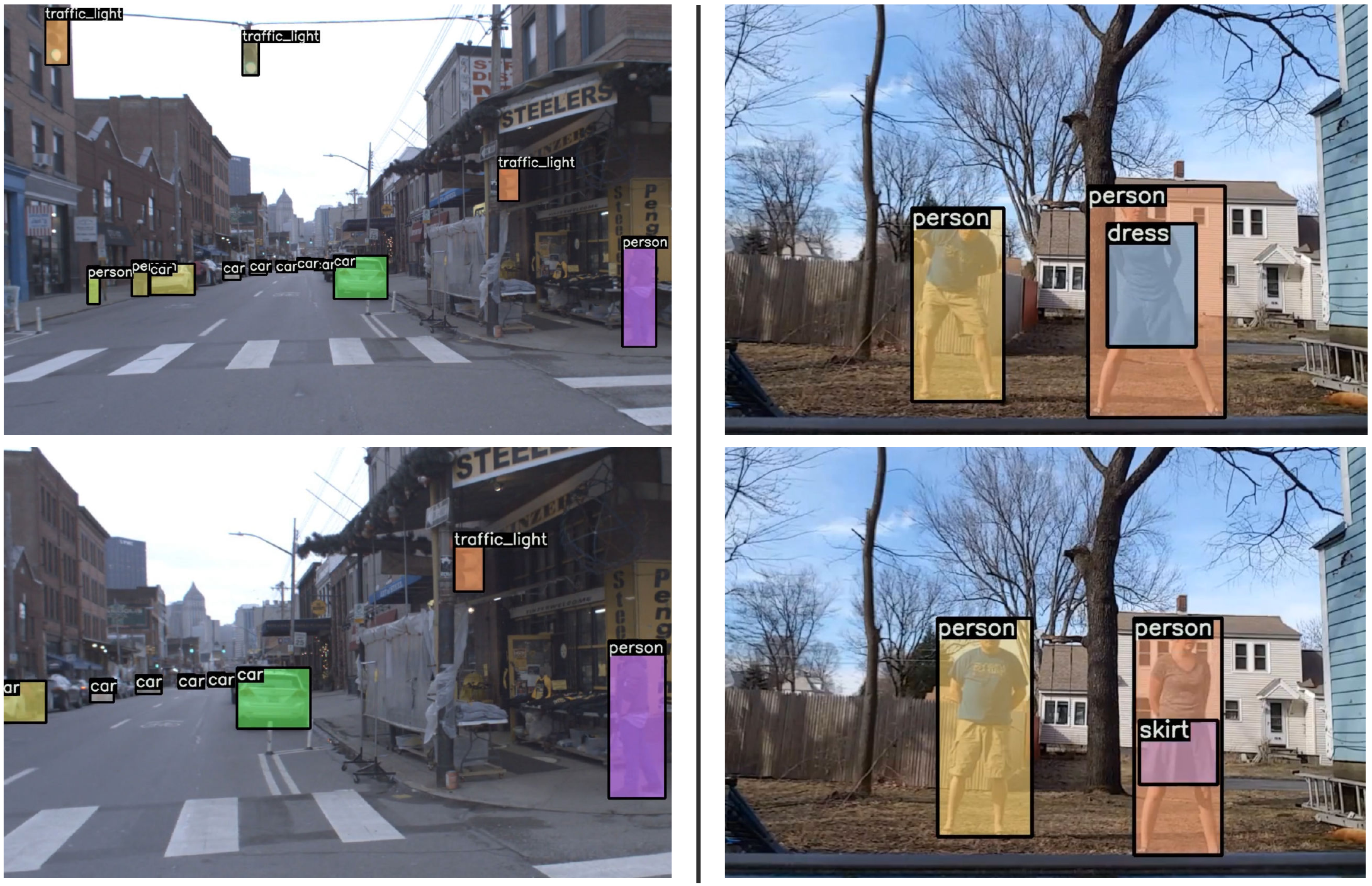}
  }{%
    \caption{SORT qualitative results, showing (left) a successful tracking result, and (right) a failure case due to semantic flicker between similar classes, suggesting that large-vocabulary tracking on \dataset{} requires additional machinery.}%
    \label{fig:sort_qualitative}
  }
\end{floatrow}
\end{figure}

\Cref{tab:mot_tao} reports results using tracking mAP on \dataset{}.
As a sanity check, we first evaluate a per-frame detector by assigning each detection to its own track.
As expected, this achieves an mAP of nearly 0 (which isn't quite 0 due to the presence of short tracks).

Next, we evaluate two multi-object tracking approaches.
We compare the SOTA Viterbi linking method to an online SORT tracker~\cite{Bewley2016_sort}.
We tune SORT hyperparameters on our diverse `train' set. Appendix~\ref{sec:appendix_tuning_sort} shows that this tuning is key for good accuracy.
The offline Viterbi algorithm takes over a month of processing time to run on our `train' set, prohibiting thorough parameter tuning.
Instead, we tune a post-processing parameter for Viterbi: the score threshold for reporting a detection at each frame.
We detail our tuning procedure in \Cref{sec:appendix_tuning}.

Surprisingly, we find that the simpler, online approach of SORT outperforms Viterbi, perhaps because the latter has been heavily tuned for \imvid{}. Because of its scalablity (to many categories and long videos) and relatively better performance, we focus on SORT for the majority of our experiments.
However, the performance of both methods remains low, suggesting \dataset{} presents a major challenge for the tracking community, requiring principled novel approaches.

To better understand the nature of the complexity of TAO, we separately measure the challenges of tracking and classification.
To this end, we first evaluate the ``track'' oracle that perfectly links per-frame detections.
It achieves a stronger mAP of 31.5, compared to 13.2 for SORT. Interestingly, providing SORT tracks with an oracle class label provides a similar improvement, boosting mAP to 30.2.
We posit that these improvements are orthogonal, and verify this by combining them; we link detections with oracle tracks and assign these tracks oracle class labels.
This provides the largest delta, dramatically improving mAP to 83.6\%. This suggests that {\em large-vocabulary tracking requires jointly improving tracking and classification accuracy (e.g., reducing semantic flicker as shown in Fig.~\ref{fig:sort_qualitative}).}

\let\oldFBaskip\FBaskip
\renewcommand\FBaskip{1em}
\setlength{\oldintextsep}{\intextsep}
\setlength\intextsep{0.5em}
\begin{wraptable}{r}{0.38\textwidth}
  \caption{Person-tracking results on \dataset{}. See text (left) for details.}
  \centering
  \resizebox{\linewidth}{!}{
  \begin{tabular}{l@{\hskip 2em}c}
    \toprule
    Method & Person AP \\\midrule
    Viterbi \cite{feichtenhofer2017detect,gkioxari2015finding} & 16.5 \\
    SORT \cite{Bewley2016_sort}                                & 18.5 \\
    Tracktor++~\cite{bergmann2019tracking}                     & 36.7 \\
    \bottomrule
  \end{tabular}
  }
  \label{tab:tao_people}
\end{wraptable}
\setlength\intextsep{\oldintextsep}
\renewcommand\FBaskip{\oldFBaskip}

\vspace{-1em}
\subsubsection{How well can we track people?}
\label{sec:results_people}
We now evaluate tracking on one particularly important category: people.
Measuring AP for individual categories in a federated dataset can be noisy~\cite{gupta2019lvis}, so we emphasize \textit{relative} performance of trackers rather than their absolute AP.
We evaluate Tracktor++~\cite{bergmann2019tracking}, the state-of-the-art method designed specifically for people tracking on our dataset, and compare it to the SORT and Viterbi baselines in Table~\ref{tab:tao_people}.
For fairness, we update Tracktor++ to use the same detector used by our SORT and Viterbi baselines, but only use the `person' predictions from this detector.
Additionally, we tune the score threshold for Tracktor++ on our `train' set, but find the method is largely robust to this parameter (see Appendix~\ref{sec:appendix_tuning_tracktor}).
We find that Tracktor++ strongly performs other approaches (36.7 AP), while SORT comes in second, modestly outperforming Viterbi (18.6 vs 16.5 AP). It is interesting to note that SORT, which can scale to all object categories, performs noticeably worse on all categories on average (13.2 mAP).
\Cref{sec:appendix_mota_person} shows that this delta between `person' and other classes is even more dramatic using the MOTA metric (6.7 overall vs 54.8 for `person').
We attribute the higher accuracy for the `person' category to two factors: (1) a rich history of focused research on this one category, which has led to more accurate detectors and trackers, and (2) more complex categories present significant challenges, such as hand-held objects which undergo repeated occlusions during interactions.

To further investigate Tracktor++'s performance, we evaluate a simpler variant of the method from~\cite{bergmann2019tracking}, which does not use appearance-based re-identification nor pixel-level frame alignment.
We evaluate this variant on \dataset{}, and find that removing these components reduces AP by over 8 points (from 36.7 to 25.9), suggesting that a majority of improvements over our baselines come from these two components.
Our results contrast those of \cite{bergmann2019tracking}, which suggest that re-id and frame alignment are not particularly helpful.
\textit{Compared to prior benchmarks, we posit the diversity of \dataset{} results in a challenging testbed for person tracking which encourages trackers robust to occlusion and camera jitter.}

\subsubsection{Do user-initialized trackers generalize better?}
\label{sec:results_user_init}
Next, we present results of recent user-initialized trackers in~\Cref{tab:compare_sot}.
For each object in \dataset{}, we provide the user-initialized tracker with a groundtruth box.
We consider two strategies for initialization. The standard approach (denoted `Init first') initializes trackers using the first frame an object appears in, and runs trackers for the rest of the video.
As the object may be partially occluded in this first frame, we additionally report a variant which initializes trackers using the frame with the largest bounding box (denoted `Init biggest'), and runs trackers forwards and backwards in time.

Unlike multi-object trackers, most user-initialized trackers report a bounding box and confidence for objects at each frame, and do not explicitly report when an object is \textit{absent}~\cite{valmadre2018long}.
To resolve this, we modify each method to report an object as absent when the confidence drops below a threshold.
We tune this threshold on the `train' set in \Cref{sec:appendix_tuning_user_init} and find that user-initialized trackers are particularly sensitive to this threshold.

\begin{table}[b!]
  \centering
  \caption{SOTA user-initialized tracking results on `val'. Surprisingly, despite using an oracle initial bounding box, these methods provide only modest improvements over a multi-object tracker. Because some user-initialized trackers are trained on videos in TAO, we re-train them on their original train set with \dataset{} videos removed, denoting this with *.}
  \begin{tabular}{lc@{\hskip 1em}c@{\hskip 1em}c@{\hskip 1em}c}
    \toprule
           & \multicolumn{2}{c}{Oracle} & \multicolumn{2}{c}{Track mAP} \\
    Method                  & Box Init & Class   & Init first & Init biggest \\\midrule
    SORT                                  &        & \cmark & \multicolumn{2}{c}{30.2} \\\midrule
    ECO  \cite{danelljan2017eco}          & \cmark & \cmark & 23.7 & 30.4 \\
    SiamMask \cite{wang2019fast}          & \cmark & \cmark & 30.8 & 37.0 \\
    SiamRPN++ LT \cite{li2019siamrpn++}   & \cmark & \cmark & 27.2 & 30.4 \\
    SiamRPN++ \cite{li2019siamrpn++}      & \cmark & \cmark & 29.7 & 35.9 \\
    ATOM* \cite{danelljan2019atom} & \cmark & \cmark & 30.9 & \textbf{38.6} \\
    DIMP* \cite{bhat2019learning} & \cmark & \cmark & \textbf{33.2} & 38.5 \\
    \bottomrule
  \end{tabular}
  \label{tab:compare_sot}
\end{table}

We compare these trackers to SORT, supplying both with a class oracle.
As expected, the use of a ground-truth initialization allows the best user-initialized methods to outperform the multi-object tracker.
However, even with an oracle box initialization and an oracle classifier, tracking remains challenging on \dataset{}.
Indeed, most user-initialized trackers provide at most modest improvements over SORT, despite using an oracle box initialization.
The `Init biggest' strategy provides stronger improvements by initializing with easier frames, but this strategy cannot be used in \textit{online} applications, as it requires access to the entire video.
\Cref{sec:appendix_mota_user_init} notes that user-initialized trackers can accurately track for a few frames after initialization, leading to improvements in MOTA, but provide little benefits in longer-term tracking.
We hypothesize that the small improvement of user-initialized trackers over SORT is due to the fact that the former are trained on videos with a small vocabulary of objects with limited occlusions, %
leading to methods that do not generalize to the most challenging cases in TAO.
One goal of user-initialized trackers is open-world tracking of objects without good detectors.
\dataset{}'s large vocabulary allows us to analyze progress towards this goal, indicating that \textit{large-vocabulary multi-object trackers may now address the open-world of objects as well as category-agnostic, user-initialized trackers.}

\section{Discussion}
Developing tracking approaches that can be deployed in-the-wild requires being able to reliably measure their performance.
With nearly 3,000 videos, \dataset{} provides such a robust evaluation benchmark.
Our analysis provides new conclusions about the state of tracking, while further raising a number of important questions to be explored in future work.

\smallsec{The role of user-initialized tracking.}
User-initialized trackers aim to track \textit{any} object, without requiring category-specific detectors.
In this work, we raise a provocative question: with the advent of large vocabulary object detectors~\cite{gupta2019lvis}, to what extent can (detection-based) multi-object trackers perform generic tracking {\em without} user initialization?
\Cref{tab:compare_sot}, for example, shows that large-vocabulary datasets (such as \dataset{} and LVIS) now allow multi-object trackers to match or outperform user-initialization for a number of categories.%

\smallsec{Specialized tracking approaches.}
Our hope in collecting \dataset{} is to measure progress in tracking in-the-wild.
A valid question is whether progress may be better achieved by building trackers for \textit{application-specific} scenarios.
An indoor robot, for example, has little need for tracking elephants.
However, success in many computer vision fields has been driven by the pursuit of \textit{generic} approaches, that can then be tailored for specific applications.
We do not build one class of object detectors for indoor scenes, and another for outdoor scenes, and yet another for surveillance videos.
We believe that tracking will similarly benefit from targeting diverse scenarios.
Of course, due to its size, \dataset{} also lends itself to use for evaluating trackers for specific scenarios or categories, as in \Cref{sec:results_people} for `person.'

\smallsec{Video object detection.}
Although image-based object detectors have shown significant improvements in recent years, our analysis in \Cref{sec:analysis_methods} suggests that simple post-processing of detection outputs remains a strong baseline for detection in videos.
While we do not emphasize it in this work, we note that \dataset{} can also be used to measure progress in video object \textit{detection}, where the goal is not to maintain the identity of objects, but simply to reliably detect them in each frame of a video.
The large vocabulary in \dataset{} particularly provides avenues for incorporating temporal information to resolve classification errors, which remain challenging (see~\Cref{fig:sort_qualitative}).

\smallsec{Acknowledgements.} We thank Jonathon Luiten and Ross Girshick for detailed feedback on the dataset and manuscript, and Nadine Chang and Kenneth Marino for reviewing early drafts. Annotations for this dataset were provided by Scale.ai. This work was supported in part by the CMU Argo AI Center for Autonomous Vehicle Research, the Inria associate team GAYA, and by the Intelligence Advanced Research Projects Activity (IARPA) via Department of Interior/Interior Business Center (DOI/IBC) contract number D17PC00345. The U.S. Government is authorized to reproduce and distribute reprints for Governmental purposes not withstanding any copyright annotation theron. Disclaimer: The views and conclusions contained herein are those of the authors and should not be interpreted as necessarily representing the official policies or endorsements, either expressed or implied of IARPA, DOI/IBC or the U.S. Government.

\clearpage

\appendix

\begin{center}\Large\bfseries Appendix\end{center}
\renewcommand{\dobib}{} %

\subfile{supplementary/supplementary_standalone.tex}

{\small
\bibliographystyle{ieee_fullname}
\bibliography{egbib}
}

\end{document}

%% file: macros.tex
\newtoggle{comments}
\toggletrue{comments}
\iftoggle{comments}{%
    \newcommand{\deva}[1]{{\leavevmode\color{blue}[Deva: #1]}}
    \newcommand{\achal}[1]{{\leavevmode\color{orange}[Achal: #1]}}
    \newcommand{\tarasha}[1]{{\leavevmode\color{magenta}[Tarasha: #1]}}
    \newcommand{\pavel}[1]{{\leavevmode\color{red}[Pavel: #1]}}
    \newcommand{\cordelia}[1]{{\leavevmode\color{purple}[Cordelia: #1]}}
}{%
  \newcommand{\deva}[1]{}
  \newcommand{\achal}[1]{}
  \newcommand{\tarasha}[1]{}
  \newcommand{\pavel}[1]{}
  \newcommand{\cordelia}[1]{}
}

\makeatletter
\newcommand{\@chapapp}{\appendixname}%
\def\authcount#1{\setcounter{auco}{#1}\setcounter{@auth}{1}}
\makeatother

\newcommand{\atmostlinewidth}{
  \ifdim\width>\columnwidth
    \columnwidth
  \else
    \width
  \fi}
\newcommand{\lightgray}[1]{\textcolor{lightgray}{#1}}

\let\origsubsubsection\subsubsection
\renewcommand{\subsubsection}[1]{\vspace{-1.5em}\noindent\origsubsubsection{#1}}
\newcommand{\smallsec}[1]{\vspace{0.2em}\noindent\textbf{#1}}
\newcommand{\dataset}{TAO}
\newcommand{\imvid}{ImageNet-Vid}

\newcommand{\TaoAllCategories}{833}
\newcommand{\TaoLvisCategories}{488}
\newcommand{\TaoNonLvisCategories}{345}
\newcommand{\TaoTrainCategories}{316}
\newcommand{\TaoEvalCategories}{785}
\newcommand{\TaoValVids}{988}
\newcommand{\TaoTrainVids}{500}
\newcommand{\TaoTestVids}{1,419}
\newcommand{\TaoEvaluationVids}{2,407}
\newcommand{\TaoTotalVids}{2,907}
\newcommand{\TaoLvisObjects}{16,144}
\newcommand{\TaoNonLvisObjects}{894}
\newcommand{\TaoLvisObjectsPerc}{95}
\newcommand{\TaoNonLvisObjectsPerc}{5}
\newcommand{\TaoAvgVidLength}{36.8}

\newcommand{\TaoAvgNumTracks}{5.9}
\newcommand{\TaoEvalTotalVidLength}{88,605}

\def\httilde{\mbox{\tt\raisebox{-.5ex}{\symbol{126}}}}

\newcommand{\cmark}{\ding{51}}%

%% file: supplementary/supplementary_standalone.tex
\Cref{sec:appendix-other} further analyzes \dataset{} annotations, including quality control and statistics.
\Cref{sec:appendix-metrics} further analyzes metrics, comparing 3D IoU to MOT challenge~\cite{milan2016mot16} metrics.
Finally, \Cref{sec:appendix_methods} further analyzes tracking methods, providing results on non-LVIS categories and hyperparameter tuning experiments.

\section{\dataset{} annotations}
\label{sec:appendix-other}

This section presents additional details about \dataset{} annotations. \Cref{sec:appendix-quality} assesses the diversity and quality of annotations. \Cref{sec:appendix_tao_stats} analyzes the size, length and motion statistics of labeled tracks. Finally, \Cref{sec:appendix_splits} provides further information regarding the construction of dataset splits.

\subsection{Annotation diversity and quality}
\label{sec:appendix-quality}
We analyze the diversity and quality of \dataset{} annotations by re-annotating 50 videos in the dataset.

\smallsec{Diversity.} One might hope that this re-annotation closely matches the original annotation.
However, in our federated setup, annotators are instructed to label only a subset of moving objects in each video. Thus, the annotations would only match if annotators had a bias towards a specific set of objects, which would hurt the diversity of \dataset{} annotations.
To verify whether this is the case, we check whether each track in the re-annotation corresponds to an object labeled in the original annotation.
Concretely, if a re-annotated track has high overlap (IoU $>0.75$) with a track in the original annotation, we assume the annotator is labeling the same object.
Our re-annotation results in 310 tracks from 50 videos. Of these 310 tracks, just over half (177, or 57\%) overlapped with those in the initial labeling with IoU $>0.75$.
The rest were \textit{new} objects not originally labeled in \dataset{}, suggesting that annotators chose to label a diverse selection of objects.

\smallsec{Quality.} Next, we evaluate the annotation agreement of the 177 re-annotated tracks that correspond to tracks originally labeled in \dataset{}.
If our annotations are of high quality, we expect these tracks to have a very high IoU (say, $>0.9$), as well as matching class labels.
Indeed, the average IoU for the 177 overlapping tracks was 0.93, indicating annotators precisely labeled the spatial and temporal extent of objects.
Finally, we evaluate the quality of the class labels in \dataset{}.
165 (93\%) were labeled with the same category as in the initial labeling;
an additional 6 (3\%) were labeled with a more precise or more general category (e.g., `jeep' vs. `car');
finally, 6 were labeled with similar labels (e.g., `kayak' vs. `canoe') or other erroneous labels.
This analysis indicates that despite the large vocabulary in \dataset{}, the class labels in \dataset{} are of high quality.

If our annotations are of high quality, we expect these tracks to have a very high IoU (say, $>0.9$), as well as matching class labels.

\smallsec{Annotation details.}
We worked closely with a professional data-labeling company, Scale.ai, to label \dataset{}.
Each track was labeled by a Scale annotator, reviewed by Scale reviewers, and finally manually inspected by the authors.

\subsection{Annotation statistics}
\label{sec:appendix_tao_stats}

\begin{figure}[t]
  \subfigure[Ratio between aspect ratio of bounding boxes between two consecutive annotated frames at 1FPS.]{
    \includegraphics[width=0.44\linewidth,trim=10 5 40 10,clip]{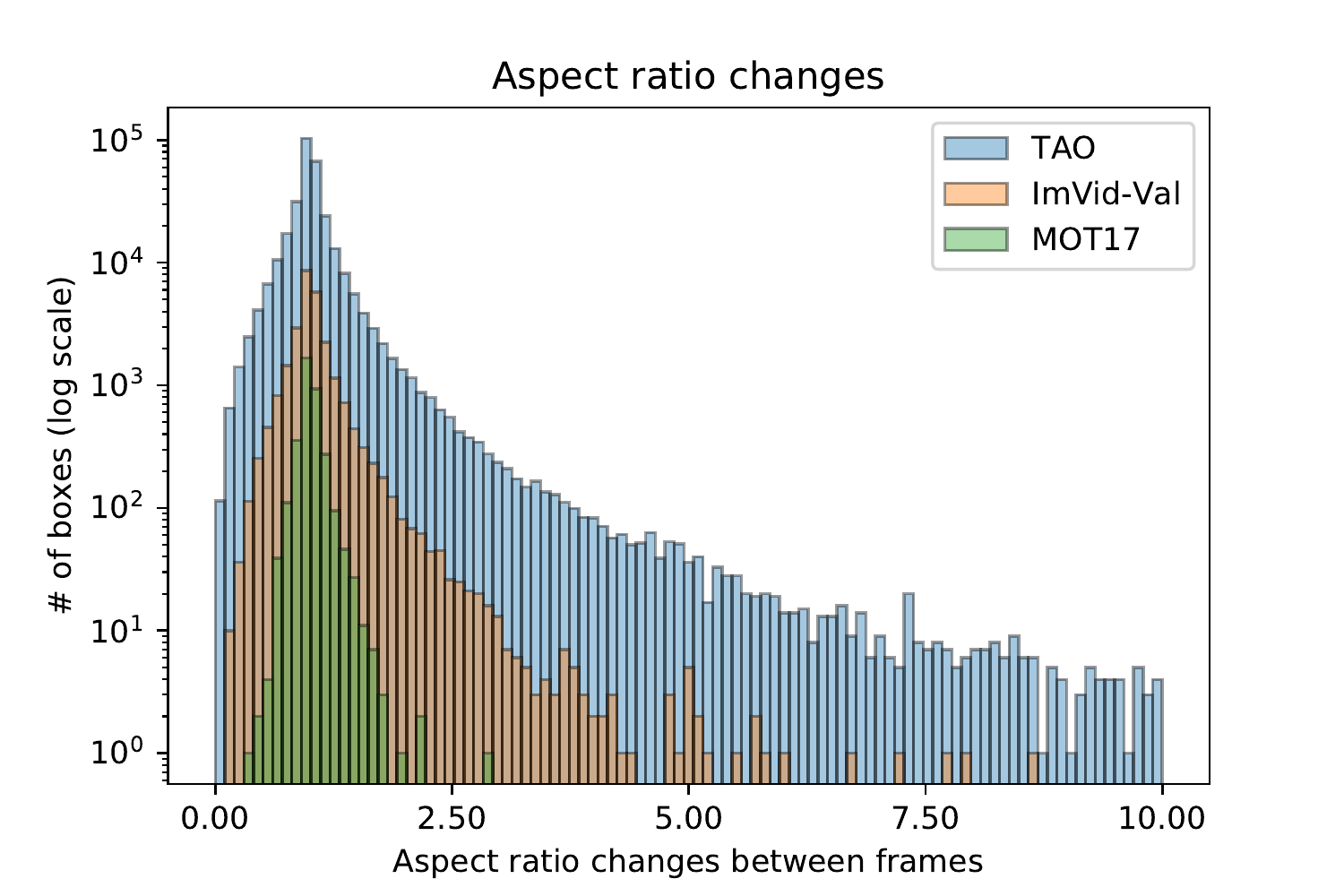}
    \label{fig:appendix_track_aspect}
  }
  \hfill
  \subfigure[Bounding box size counts relative to size of the image.]{
    \includegraphics[width=0.44\linewidth,trim=10 5 40 10,clip]{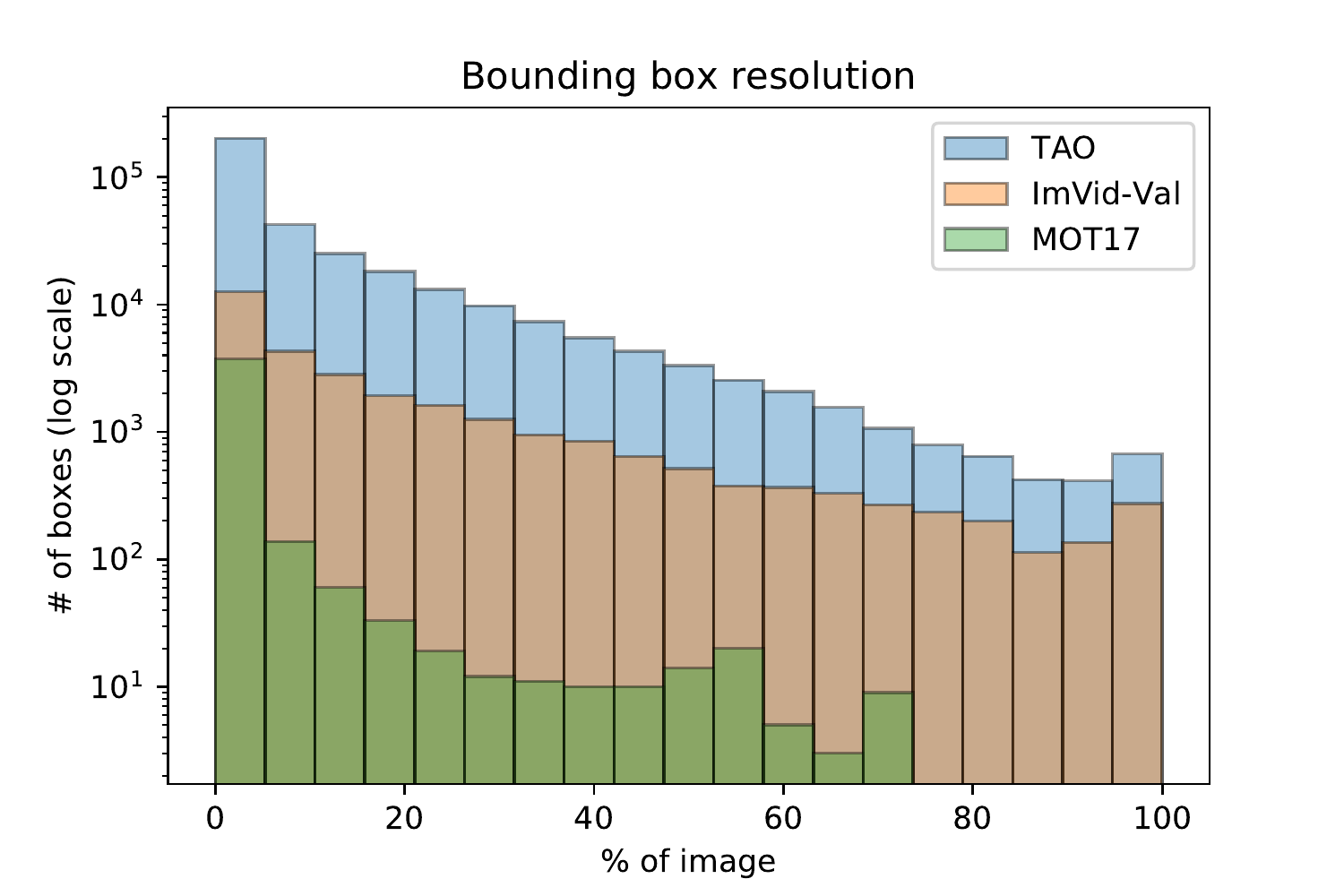}
    \label{fig:appendix_track_res}
  }
  \subfigure[Distance between center of objects between two consecutive annotated frames at 1FPS.]{
    \includegraphics[width=0.45\linewidth,trim=10 5 40 10,clip]{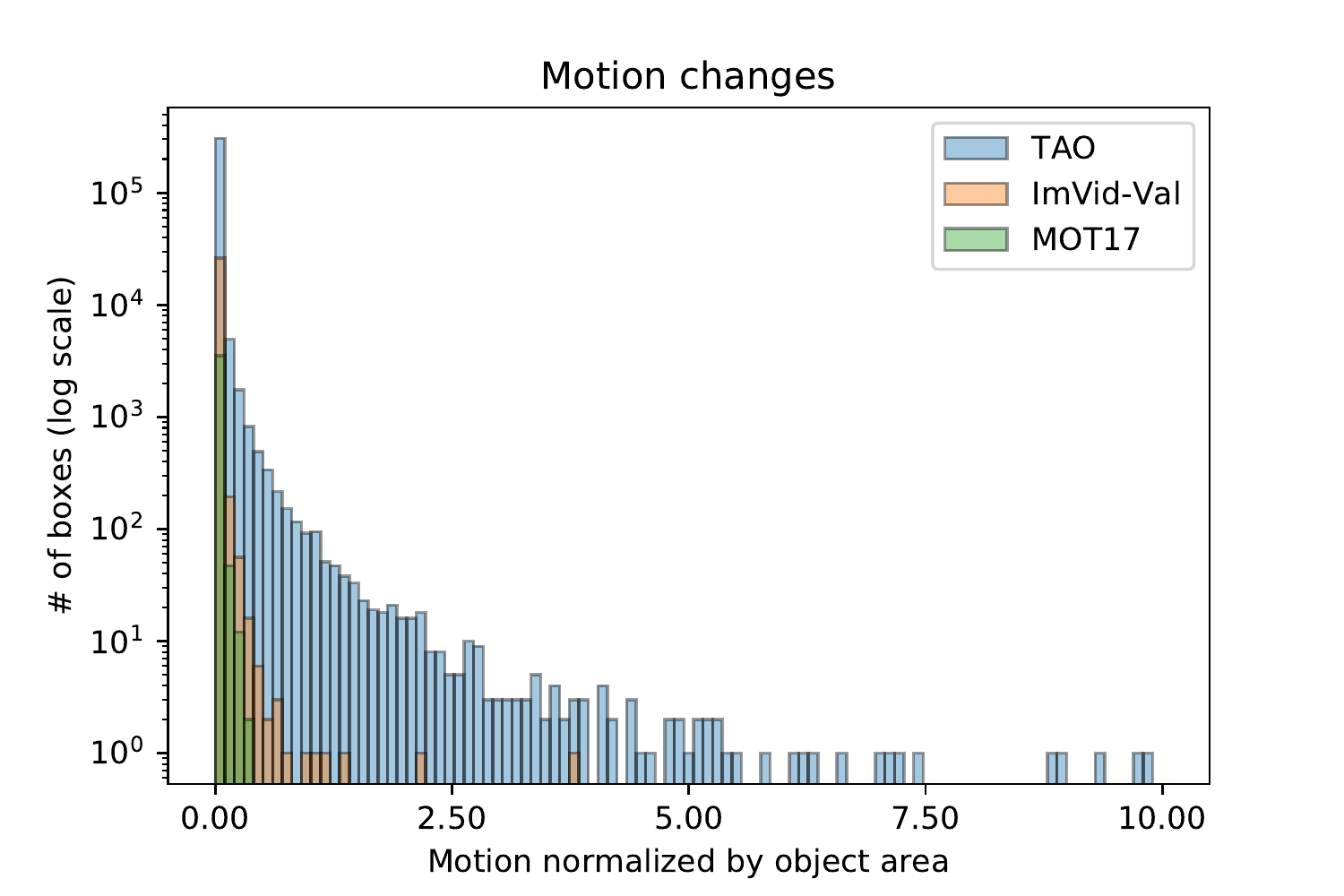}
    \label{fig:appendix_track_motions}
  }
  \hfill
  \subfigure[Track length counts, relative to video lengths.]{
    \includegraphics[width=0.45\linewidth]{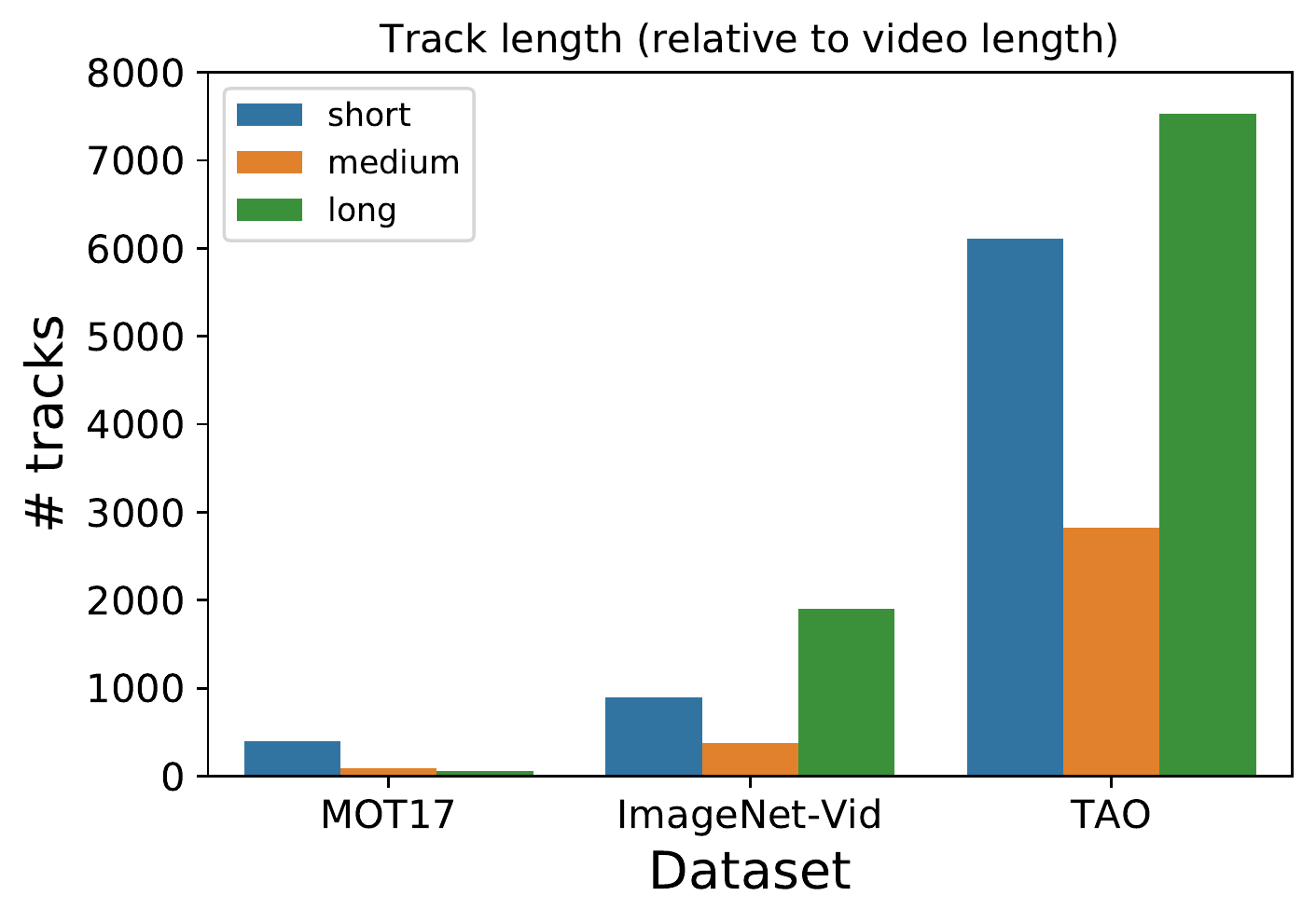}
    \label{fig:appendix_track_lengths}
  }
  \caption{Additional statistics of the \dataset{} dataset. See \Cref{sec:appendix_tao_stats} for details.}
  \label{fig:appendix_tao_plots}
\end{figure}

We present further analysis of the annotated tracks in \dataset{} in~\Cref{fig:appendix_tao_plots}.
We compare \dataset{} to MOT-17~\cite{milan2016mot16} and ImageNet-Vid~\cite{russakovsky2015imagenet}, which are benchmark datasets where the Viterbi~\cite{gkioxari2015finding,feichtenhofer2017detect} and the Tracktor~\cite{bergmann2019tracking} approaches were originally evaluated.

\Cref{fig:appendix_track_aspect} shows the distribution of changes in aspect ratio between two annotated frames at 1FPS.
Concretely, the aspect ratio change is $(w_t / h_t) / (w_{t-1} / h_{t-1})$, where $w_t, h_t$ are the width and height of the object at time $t$, respectively (see \cite{kristan2016novel}).
This metric can be used to understand the types of motion in tracking datasets.
MOT-17 focuses on people, which largely have the same aspect ratio over time.
ImageNet-Vid has a slightly more diverse distribution of changes in aspect ratio, but \dataset{} has by far the most diverse distribution, due to its large size and diversity of categories.

\Cref{fig:appendix_track_res} plots the distribution of bounding box resolution as a percentage of the image.
MOT-17 tends to have smaller bounding boxes, while \dataset{} and ImageNet-Vid have a variety of object sizes.
Note again that \dataset{} presents a much larger number of tracks used for evaluation, visible even on the log-scale in \Cref{fig:appendix_track_res}, than ImageNet-Vid val.

\Cref{fig:appendix_track_motions} presents the distribution of object motion, proportional to the size of the object.
Concretely, let $a_t$ be the area of the bounding box at time $t$.
We define the distance in $x$ as
$d_t^x = \frac{\lVert x_t - x_{t-1} \rVert}{a_{t-1}}$, and similarly for $d_t^y$.
Then,
$d_t = \lVert [d_t^x, d_t^y] \rVert_2^2$.
As with~\Cref{fig:appendix_track_aspect}, we plot these changes at 1FPS so that the annotation rate does not impact the plot.
We note that \dataset{} contains a variety of object motions, including extremely fast motions for small objects, as evidenced by the number of boxes with motion change larger than 5.0.

\Cref{fig:appendix_track_lengths} shows the distribution of object track lengths in~\dataset{}. For clarity, we group the tracks into 3 bins based on length: short, medium and long, which correspond to  less than 1/3, between  1/3 and 2/3, and greater than 2/3 of the length of the video.
The plot shows that \dataset{} provides diversity in object track length, requiring methods to be able to track for long periods of time, while also being able to recognize when an object is missing.
By contrast, MOT-17 is biased towards short tracks, while ImageNet-Vid is biased towards long tracks.

Finally, we present statistics of four recent benchmarks for user-initialized
tracking (or single-object tracking) in \Cref{tab:appendix_sot_dataset_stats}.
We note that datasets tend to benchmark tracking on a smaller number of categories than~\dataset{}, and on far fewer videos.
While this may be appealing from a computational perspective, we argue that progress in tracking requires evaluating on a large, diverse set of scenarios, ensuring that methods do not overfit to any small set of videos or environments.
Further, unlike standard user-initialized tracking datasets, \dataset{} contains nearly 5x as many tracks per video, leading to a much larger number of total tracks compared to prior benchmarks.

\begin{table*}
  \centering
  \caption{Statistics of major user-initialized tracking datasets.}
   \resizebox{\atmostlinewidth}{!}{
  \begin{tabular}{lccccccccc}
    \toprule
    Dataset &
    \multicolumn{2}{c}{
      \begin{tabular}{cc}
        \multicolumn{2}{c}{Classes} \\
        Eval. & Train
      \end{tabular}
    } &
    \multicolumn{2}{c}{
      \begin{tabular}{cc}
        \multicolumn{2}{c}{Videos} \\
        Eval. & Train
      \end{tabular}
    } &
    \begin{tabular}{c}Avg\\length (s)\end{tabular} &
    \begin{tabular}{c}Tracks \\ / video\end{tabular} &
    \begin{tabular}{c}Min \\resolution\end{tabular} &
    \begin{tabular}{c}Ann.\\fps\end{tabular} &
    \begin{tabular}{c}Total Eval\\length (s)\end{tabular} \\\midrule
    GOT-10k~\cite{huang2018got}\footnote{Stats from the GOT-10k dataset release, which differ from those in~\cite{huang2018got}.}    & 84 & 480 & 360 & 9,335 & 12.2   & 1    & 270x480 & 10 & 4,384 \\
    OxUvA~\cite{valmadre2018long}  & 22 & 0 & 366 & 0 & 141.2   & 1.1  & 192x144 & 1 & 51,667 \\
    LaSOT~\cite{fan2019lasot}      & 70 & 70 & 280 & 1,120 & 82.1    & 1    & 202x360 & \httilde 25 & 23,520 \\
    TrackingNet~\cite{muller2018trackingnet}
                                   & 27 & 27 & 511 & 30,132 & 14.7  & 1    &   270x360     & \httilde 28 & 7,511 \\
    \dataset{} (Ours)\footnote{\dataset{} train and eval contain partially overlapping subsets of the overall \TaoAllCategories{} categories.} & \TaoEvalCategories & \TaoTrainCategories{} & \TaoEvaluationVids{} & \TaoTrainVids & \TaoAvgVidLength{}
      & \TaoAvgNumTracks{} & 640x480 & 1 & \TaoEvalTotalVidLength{} \\
    \bottomrule
  \end{tabular}
   }
  \label{tab:appendix_sot_dataset_stats}
\end{table*}

\subsection{Split construction}
\label{sec:appendix_splits}

We construct our `train', `val', and `test' splits to respect the following constraints:
\begin{itemize}
  \itemsep0em
  \item \textbf{Charades} contains videos recorded by mechanical turk workers, and one worker may contribute multiple videos to Charades. We ensure that any two videos uploaded by the same worker falls in the same split.
  \item \textbf{ArgoVerse} contains video recordings from different cameras from the same driving sequence. We ensure that all videos from the same driving sequence fall in the same split.
  \item \textbf{HACS} contains videos uploaded to YouTube. Any two videos uploaded by the same YouTube user, or uploaded to the same YouTube channel, must fall in the same split.
  \item \textbf{AVA.} We split AVA movies into multiple contiguous shots, and ensure shots from the same movie fall in the same split.
  \item \textbf{YFCC100M} contains videos uploaded to Flickr. Any two videos uploaded by the same Flickr user fall in the same split.
  \item \textbf{BDD} and \textbf{LaSOT}: No constraints are applied for split construction.
\end{itemize}

\section{Metrics}
\label{sec:appendix-metrics}

In this section, we further analyze the 3D IoU metric (\ref{sec:appendix-metric-discussion}), report results using the MOT challenge~\cite{milan2016mot16} metrics (\ref{sec:appendix_mota}), and finally present per-category APs for SORT (\ref{sec:appendix-sort-per-cat}).

\subsection{3D IoU Discussion}
\label{sec:appendix-metric-discussion}

\begin{figure}
    \centering
    \includegraphics[width=0.5\linewidth]{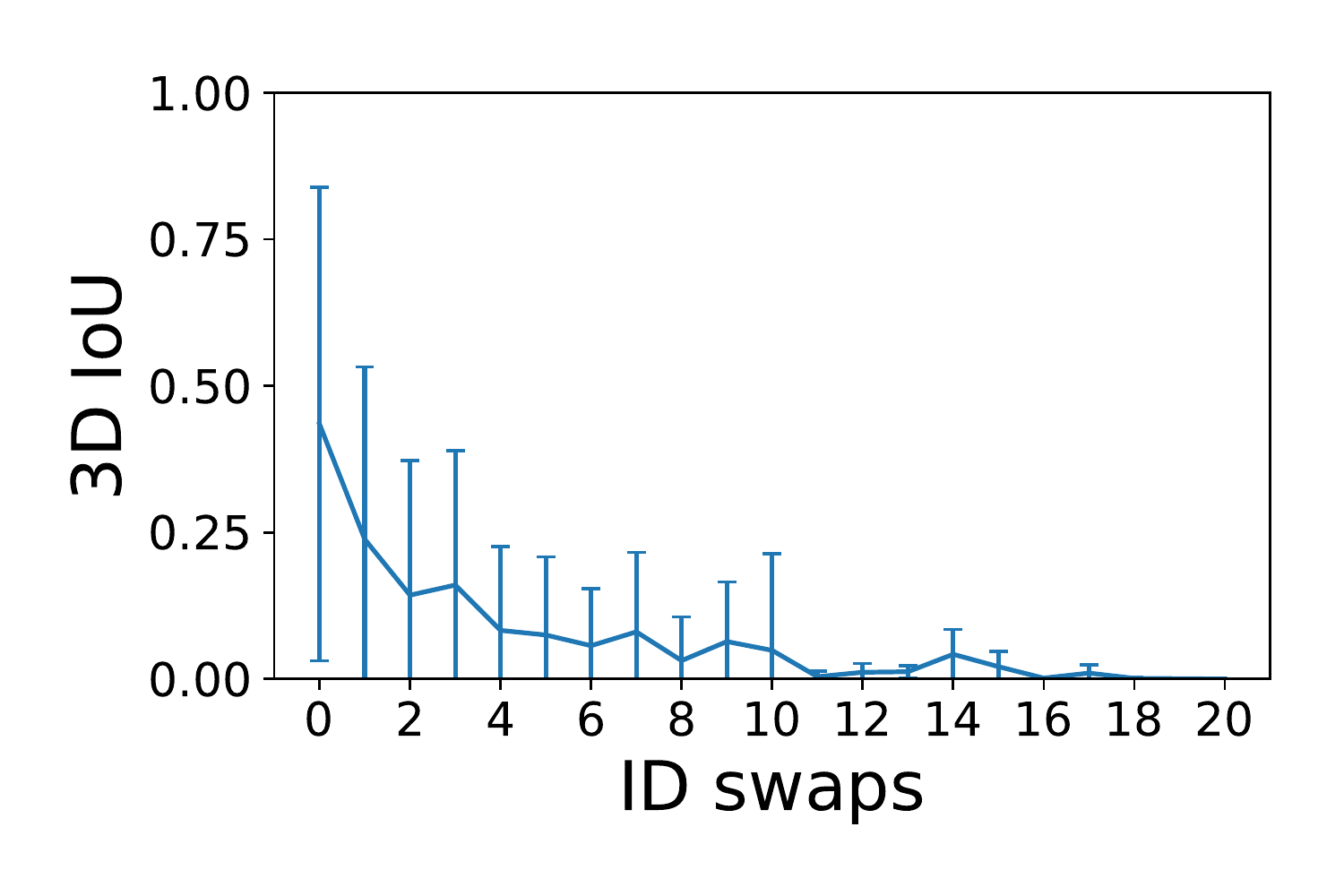}
    \caption{For each pair of predicted and groundtruth tracks matched to each other on \dataset{}, we compute the 3D IoU and number of ID swaps. Above, we plot the mean and variance of 3D IoU vs. ID swaps across tracks, and show that 3D IoU drops as the number of ID swaps increases.}
    \label{fig:appendix_iou_vs_idswaps}
\end{figure}

The mAP metric using 3D IoU provides a concise, interpretable evaluation of tracking in the wild, as evidenced by its use in recent datasets for multi-object tracking with many categories~\cite{deng2009imagenet,yang2019video}.
We further discuss this metric below:

\smallsec{Relation to identity swaps.} \Cref{fig:appendix_iou_vs_idswaps} shows that 3D IoU is correlated with a key metric for tracking: identity swaps, as measured by the MOT challenge~\cite{milan2016mot16} metrics.

\smallsec{Partial credit.} Evaluating trackers with mAP requires specifying an IoU threshold, which we set to $0.5$ throughout the experiments in the main paper.
Consequentially, trackers do not receive \textit{partial credit} for tracking an object for short time periods.
Consider two trackers: Tracker A perfectly tracks an object for 30\% of its track length, while Tracker B only tracks the object 5\% of the time.
At an IoU threshold of 0.5, A and B will result in the same mAP.
By contrast, metrics such as MOTA and ID-F1 will be significantly higher for A than for B.
The 3D IoU mAP metric takes inspiration from image-based detection metrics: as object detectors receive no credit for loose localizations, object trackers receive no credit for loosely tracking objects for a few frames.
If desired, the mAP metric can be modified to provide partial credit by averaging over multiple IoU thresholds, similar to the COCO evaluation~\cite{lin2014microsoft}.

\smallsec{Confidence estimates.} Metrics such as MOTA~\cite{bernardin2008evaluating} and ID-F1~\cite{wu2006tracking} metrics do not evaluate the confidence provided by many modern tracking approaches.
By contrast, our mAP metric evaluates these explicitly when tracing out the precision-recall curve.
This allows us to evaluate methods across diverse application scenarios, which may have different tradeoffs between precision and recall.

\smallsec{Impact of object size.} 3D IoU is computed over spatio-temporal volumes.
As such, frames where an object's bounding box is \textit{large} have a greater impact on the spatio-temporal volume than frames where an object's bounding box is small, thus factoring in more heavily into the IoU measure.
We note that for many applications, such as navigation, this is a desirable property, as accurate localization and tracking is more important for nearby objects.
For other applications, additional diagnostics, such as MOTA (\Cref{sec:appendix_mota}), can be used for further analysis.

\subsection{MOTA results}
\label{sec:appendix_mota}

For completeness, we present results using the MOT challenge suite of metrics~\cite{milan2016mot16}:
MOTA~\cite{bernardin2008evaluating}, ID-F1~\cite{ristani2016performance}, mostly-tracked (MT) tracks and mostly-lost (ML) tracks~\cite{wu2006tracking}, false-positives (FP), false-negatives (FN) and identity swaps (ID Sw.), computed using the \texttt{py-motmetrics} library~\cite{cheind2020pymot}.
To do this, we first make two modifications to the MOT metrics:

\textbf{Federated MOTA and ID-F1.} We update the MOTA and ID-F1 metrics for a federated dataset by only counting false positives (FPs) for a category $c$ in video $v$ if we know that all instances of category $c$ are annotated in video $v$ (i.e., if $v$ is in $P_c$ or $N_c$ as defined in Sec. 3 of our paper).
While this approach is not perfect, as it can over-estimate the performance of a tracker, it provides a simple adaptation to the federated setup.

\textbf{Multiple categories.} The MOT metrics are usually reported for a single category~\cite{milan2016mot16}, or separately for a small number categories~\cite{geiger2013vision}.
This is not a scalable strategy for \dataset{}, which contains \TaoAllCategories{} categories.
Instead, we compute metrics separately per category, and combine them across categories.
Concretely, for metrics such as MOTA and ID-F1, we report the average value across categories.
For counters, including MT (mostly-tracked), ML (mostly-lost), FP (false-positives), FN (false-negatives) and ID Sw. (identity switches), we report the sum across categories.
Note that while MOTA and ID-F1 are balanced across categories, the `counters' are heavily dominated by the most frequent categories.

\textbf{Thresholds.} Unlike mAP, the MOT metrics require picking a confidence threshold for evaluation.
To do this, we search over track score thresholds on \dataset{} train and report results in \Cref{tab:appendix_mota_tuning}.
For Viterbi and user-initialized trackers, the track score threshold is applied after the tracker per-frame score threshold tuned in \Cref{sec:appendix_tuning}.
Hence, the MOTA for track thresholds below the per-frame threshold are equivalent (e.g., for DIMP, the optimal per-frame threshold is 0.5, and so the MOTA for thresholds below 0.5 is exactly the same: 22.7).

\addtolength{\tabcolsep}{2pt}
\begin{table}[t]
  \caption{Results from tuning track score thresholds for multi-object trackers, user-initialized trackers, and Tracktor++ on \dataset{} train, reporting MOTA.}
\centering
\begin{tabular}{@{}lcccccccccc}
  \toprule
  Tracker   & 0.1  & 0.2  & 0.3  & 0.4  & 0.5  & 0.6  & 0.7  & 0.8  & 0.9 & 1.0 \\ \midrule
  Detector  & -18.1 & -9.7 & -6.1 & -3.8 & -2.2 & -1.3 & -1.0 & -0.3 & -0.01 & \textbf{0.0} \\\midrule
  SORT      & -3.0 & 7.7 & 7.7 & 7.9 & \textbf{8.5} & 6.9 & 5.4 & 3.6 & 2.5 & 0.0 \\
  Viterbi   & -8.4 & 2.5 & 5.4 & 5.6 & 6.2 & \textbf{6.8} & 5.3 & 5.3 & 3.3 & 0.0 \\\midrule
  ATOM     & 21.8 & 21.8 & 21.8 & 21.8 & 21.8 & 21.8 & \textbf{27.2} & 19.8 & 8.2 & 0.0 \\
  DIMP     & 22.7 & 22.7 & 22.7 & 22.7 & 22.7 & \textbf{22.7} & 22.6 & 21.4 & 20.3 & 19.1 \\
  ECO      & 0.7  & 0.7  & 0.3  & 1.5  & 7.0  & 8.1  & \textbf{12.6} & 6.1  & 0.3 & 0.0  \\
  SiamMask & 19.7 & 19.7 & 19.7 & 19.7 & 19.7 & 19.7 & 19.7 & 19.7 & \textbf{19.9} & 0.0 \\
  SiamRPN++ & 21.0 & 21.0 & 21.0 & 21.0 & 21.0 & 21.0 & 21.0 & 21.0 & \textbf{25.5} & 0.0 \\
  SiamRPN++ LT & 22.9 & 22.9 & 22.9 & 22.9 & 22.9 & 22.9 & 22.9 & 22.9 & \textbf{22.9} & 0.0 \\
  \bottomrule
  \multicolumn{10}{c}{Person-only evaluation} \\
  \toprule
  Tracktor++      & 65.9 & 66.0 & 66.2 & 66.5 & 67.2 & 67.9 & \textbf{68.4} & 67.8 & 63.0 \\\bottomrule
  \end{tabular}
  \label{tab:appendix_mota_tuning}
\end{table}
\addtolength{\tabcolsep}{-2pt}

We use the optimal thresholds from the train set to report results on the validation set for multi-object trackers in \Cref{tab:appendix_mota_results}, for user-initialized trackers in \Cref{tab:appendix_mota_sot}, and for person-tracking in \Cref{tab:appendix_mota_people}.
In general, we find that the conclusions drawn in our main paper using mAP are consistent with experiments using MOTA, with two exceptions.

\addtolength{\tabcolsep}{2pt}
\begin{table}[H]
  \caption{MOT challenge metrics for multi-object trackers on \dataset{} validation. As the `Track' oracle implicitly removes false positive detections, we set score thresholds to 0 when it is used.}
  \centering
  \resizebox{\atmostlinewidth}{!}{
  \begin{tabular}{lccccccccc}
    \toprule
          & \multicolumn{2}{c}{Oracle} &&&&&&& \\
    Method & Class & Track &
    MOTA $\uparrow$ & ID-F1 $\uparrow$ & MT $\uparrow$ & ML $\downarrow$
     & FP $\downarrow$ & FN $\downarrow$ & ID Sw. $\downarrow$ \\
    \midrule
    Detection  & & &
     -2.3 & 1.3 & 1,495 & 1,941 & 3,492 & 60,776 & 48,377 \\\midrule
    Viterbi & & &
      5.6 & 10.0 & 1,407 & 2,409 & 5,367 & 62,341 & 10,262 \\
    SORT & & &
      6.7 & 10.4 & 1,687 & 2,117 & 4,146 & 59,481 & 4,772 \\
    Detection  & & \cmark &
      38.8 & 48.4 & 2,191 & 919 & 0 & 42,796 & 0 \\\midrule
    Viterbi  & \cmark & &
      8.3 & 13.8 & 1447 & 2361 & 5595 & 60787 & 10292 \\
    SORT  & \cmark & &
      11.3 & 15.6 & 1,725 & 2,066 & 4,165 & 58,418 & 4,773 \\
    Detection  & \cmark & \cmark &
      83.2 & 89.6 & 3,806 & 188 & 0 & 17018 & 6 \\
    \bottomrule
  \end{tabular}
  }
  \label{tab:appendix_mota_results}
\end{table}
\addtolength{\tabcolsep}{-2pt}

\subsubsection{User init.}
\label{sec:appendix_mota_user_init}
First, \Cref{tab:appendix_mota_sot} shows that user-initialized trackers provide significant improvements over SORT using MOTA and ID-F1, while this did not hold for mAP. These metrics provide partial credit for tracking objects for short periods of time, while mAP (with an 3D IoU threshold of 0.5) requires tracking an object for at least half its track length (see~\Cref{sec:appendix-metric-discussion}).
One can obtain mAP rankings consistent with MOTA/ID-F1 by using an artificially low IoU threshold; at a threshold of 0.1, DIMP strongly outperforms SORT, 71.0 mAP to 36.9 mAP. These results reinforce the notion that user-initialization is helpful for tracking short periods after initialization, but less helpful in the long term.

\addtolength{\tabcolsep}{2pt}
\begin{table}[H]
  \caption{MOT challenge metrics on \dataset{} validation, comparing user-initialized trackers with SORT using a class oracle.}
  \centering
  \resizebox{\atmostlinewidth}{!}{
  \begin{tabular}{lccccccccc}
    \toprule
    & \multicolumn{2}{c}{Oracle} & &&&&&& \\
    Method & Box Init & Class &
     MOTA $\uparrow$ & ID-F1 $\uparrow$ & MT $\uparrow$ & ML $\downarrow$
     & FP $\downarrow$ & FN $\downarrow$ & ID Sw. $\downarrow$ \\
    \midrule
    SORT & & \cmark &
      11.3 & 15.6 & 1,725 & 2,066 & 4,165 & 58,418 & 4,773 \\
    \midrule
    ECO & \cmark & \cmark &
      11.8 & 24.0 & 753 & 4341 & 5395 & 85415 & 42 \\
    SiamRPN++ LT & \cmark & \cmark &
      13.1 & 54.0 & 2,292 & 753 & 19282 & 42255 & 2103 \\
    SiamRPN++ & \cmark & \cmark &
      14.6 & 49.9 & 2,110 & 1229 & 16630 & 45612 & 1411 \\
    ATOM & \cmark & \cmark &
      16.9 & 46.7 & 1,694 & 2,274 & 14,625 & 55,875 & 481 \\
    DIMP & \cmark & \cmark &
      24.4 & 55.1 & 2,279 & 870 & 16,966 & 42,729 & 1,290 \\
      \bottomrule
  \end{tabular}
  }
  \label{tab:appendix_mota_sot}
\end{table}
\addtolength{\tabcolsep}{-2pt}

\subsubsection{MOTA-Person.}
\label{sec:appendix_mota_person}
Second, as noted in the main paper, \Cref{tab:appendix_mota_people} shows that MOTA-person is significantly higher than MOTA-overall (6.7 vs 54.8 for SORT), whereas the delta is smaller under mAP (13.2 vs 18.5 for SORT). We find MOT metrics heavily reward accurate detection while 3D IoU heavily penalizes inaccurate tracking. Because person detectors strongly outperform other category detectors on average, this is manifested as a high MOTA-person score.

\addtolength{\tabcolsep}{2pt}
\begin{table}[H]
  \caption{MOT challenge metrics on \dataset{} validation for the `person' category.}
  \centering
  \begin{tabular}{lcccccccc}
    \toprule
    Method &
     MOTA $\uparrow$ & ID-F1 $\uparrow$ & MT $\uparrow$ & ML $\downarrow$
     & FP $\downarrow$ & FN $\downarrow$ & ID Sw. $\downarrow$ \\
    \midrule
    Viterbi
      & 44.5 & 50.4 & 939 & 741 & 21,678 & 3,167 & 7,128 \\
    SORT
      & 54.8 & 56.2 & 1,078 & 542 & 20,025 & 2,432 & 3,567 \\
    Tracktor++
      & 66.6 & 64.8 & 1,529 & 411 & 12,910 & 2,821 & 3,487 \\
      \bottomrule
  \end{tabular}
  \label{tab:appendix_mota_people}
\end{table}
\addtolength{\tabcolsep}{-2pt}

\textbf{Other benchmarks.}
Finally, we directly compare Tracktor++ on \dataset{} with its performance on the MOT-17 dataset.
\Cref{tab:appendix_tracktor_mota} shows that the more sophisticated components of Tracktor++ (re-identification and motion compensation) lead to significant improvements on TAO, suggesting \dataset{}  encourages trackers robust to common tracking challenges, including occlusion and camera motion.

\begin{table}
  \centering
  \caption{MOTA on \dataset{} val vs. MOT-17, for Tracktor. \dataset{} encourages trackers robust to camera motion and occlusion, as noted by the significant improvement to Tracktor using the reID and camera motion compensation (CMC) components.}
  \begin{tabular}{lcc@{\hskip 0.5cm}cc}
    \toprule
           & \multicolumn{2}{c}{\dataset} & \multicolumn{2}{c}{MOT-17} \\
    Method & train & val & train & test \\\midrule
    Tracktor
      & 63.8 & 61.6    & 61.5 & - \\
    Tracktor++ (reID + CMC)
      & 68.4 & 66.6 & 61.9 & 53.5 \\
    \bottomrule
  \end{tabular}
  \label{tab:appendix_tracktor_mota}
\end{table}

\subsection{AP per category}
\label{sec:appendix-sort-per-cat}
\begin{figure}
    \centering
    \includegraphics[width=\textwidth,height=0.9\textheight,keepaspectratio]{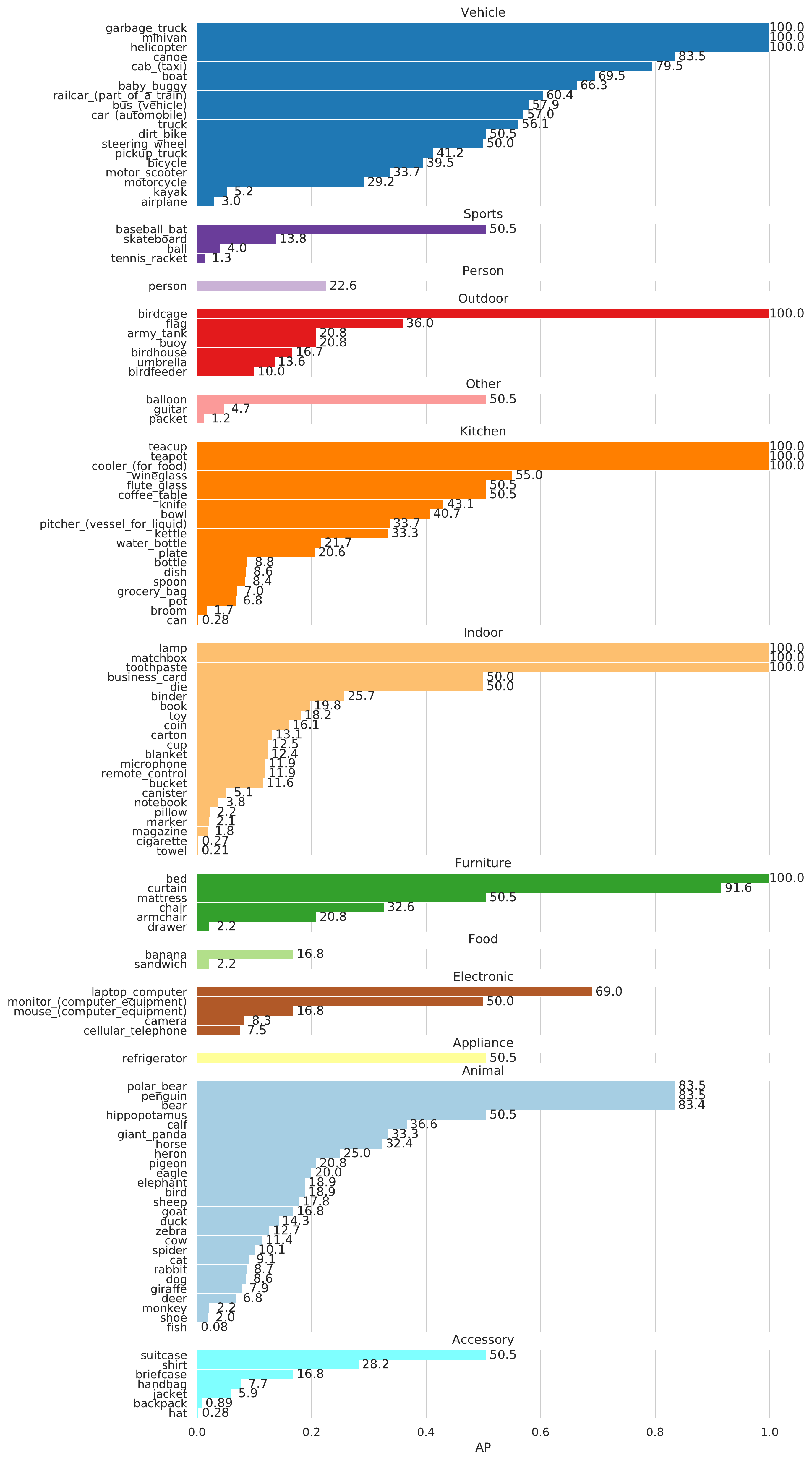}
    \caption{Per-category AP for the SORT algorithm, omitting 180 categories which result in zero AP for conciseness. As common in large-vocabulary datasets (LVIS, ADE-20K, LabelMe), average accuracy is dominated by classes in the tail, many of which result in 0 AP. Note that AP for individual categories can be noisy in a federated setup~\cite{gupta2019lvis}.} %
    \label{fig:appendix_category_aps}
\end{figure}

We present per-category APs in \Cref{fig:appendix_category_aps} for the SORT algorithm reported in the main paper, though we note that AP for individual categories can be noisy in a federated setup~\cite{gupta2019lvis}.
Note that for 180 categories, this algorithm achieves 0 AP; for conciseness, we plot only the categories with non-zero AP.

\section{Additional tracking results}
\label{sec:appendix_methods}

\Cref{sec:appendix-all-categories} presents results for user-initialized trackers on \textit{all} categories in \dataset{}.
\Cref{sec:appendix_tuning} reports results from tuning trackers on \dataset{} train.

\subsection{User-initialized trackers on all categories}
\label{sec:appendix-all-categories}

In the main paper, we focus our analysis on a subset of \dataset{} categories which exist in the LVIS~\cite{gupta2019lvis} dataset, allowing us to repurpose existing object detectors for multi-object tracking. Here, we evaluate user-initialized trackers (which do not require object detectors) on the remaining categories in \dataset{} (\Cref{tab:appendix_sot_non_lvis}).
We generally find that the results are consistent with the results on the LVIS categories.

\begin{table}
  \centering
  \caption{Results on non-LVIS, free-form text categories in \dataset{} validation.}
  \begin{tabular}{l@{\hskip 0.2cm}c}
    \toprule
    Method & Non-LVIS categories, validation \\\midrule
    ECO          & 24.1 \\
    SiamMask     & 27.0 \\
    SiamRPN++    & 27.7 \\
    SiamRPN++ LT & 25.1 \\
    ATOM         & 29.5 \\
    DIMP         & \textbf{29.6} \\
    \bottomrule
  \end{tabular}
  \label{tab:appendix_sot_non_lvis}
\end{table}

\subsection{Hyperparameter tuning}
\label{sec:appendix_tuning}

This section reports detailed results of tuning each tracker on \dataset{} train, as well as information about the detector used for SORT, Viterbi and Tracktor++ (\ref{sec:appendix_detector}).

{\bf Preliminary: Score thresholds.}
Before discussing the details of each tracker, we define three different score thresholds used by trackers, and refer to them by name throughout the appendix:

\begin{enumerate}
  \item Detection score: This is the confidence reported by a detector for each object at each frame, \textit{before} any tracking has taken place.
  \item Tracker per-frame score: This is the confidence reported by the tracker for each object at each frame, \textit{after} tracking is complete.
  \item Track score: This is the confidence reported by the tracker for each object \textit{track} throughout the video. This confidence is used to rank tracks when computing mAP. When computing MOTA, we tune the threshold for reporting tracks using the track score, as described in \Cref{sec:appendix_mota}.
\end{enumerate}

\subsubsection{SORT.}
\label{sec:appendix_tuning_sort}
We tune three parameters internal to SORT, as well as parameters of the underlying detector in \Cref{tab:appendix_sort_tuning}.
We tune the following SORT parameters:
\begin{enumerate}
  \item Det / image: Max number of detections output by the detector per image.
  \item Detection score
  \item \texttt{max\_age}: How many frames tracks are kept `alive' for, without any detections being matched to them.
  \item \texttt{min\_hits}: How many frames a track must be alive for before it is considered `confirmed' and output.
  \item \texttt{min\_iou}: Minimum IoU between a track and a detection required for linking the two.
  \item NMS Thresh: The NMS IoU threshold used by the detector. We experiment with more aggressive NMS, which may make the task of linking detections using IoU easier.
\end{enumerate}

The first row in~\Cref{tab:appendix_sort_tuning} corresponds to the default SORT parameters.
Due to the significant motion and long duration of sequences in \dataset{} (see \Cref{sec:appendix_tao_stats}), we find that increasing \texttt{max\_age} and decreasing \texttt{min\_iou} and \texttt{min\_hits} helps significantly with accuracy.
Additionally, we find that outputting more boxes per image consistently improves accuracy.
Lowering the score threshold from 0.1 to 0.0005 results in a 2.1 point improvement from 8.2 to 11.3, and lowering the NMS and score thresholds provides even more significant improvements, from 11.3 to 16.3.

\addtolength{\tabcolsep}{3pt}
\begin{table}[t]
  \caption{Results from tuning SORT parameters (by coordinate descent) on \dataset{} train, where the active coordinate (parameter) is highlighted.}
  \centering
  \begin{tabular}{cccccc@{\hskip 0.5cm}c}
    \toprule
    \multicolumn{6}{c}{Params} & \\
    NMS Thresh & Det / image & Det score & \texttt{max\_age} & \texttt{min\_hits} & \texttt{min\_iou} & Track mAP \\
    \midrule
    0.5 & 300 & 0.1    & 1 & 3 & 0.3 & 4.3 \\\midrule
    \lightgray{0.5} & \lightgray{300} & \lightgray{0.1}    & \lightgray{1} & \lightgray{3} & 0.1 & 5.0 \\
    \lightgray{0.5} & \lightgray{300} & \lightgray{0.1}    & \lightgray{1} & \lightgray{3} & 0.5 & 4.3 \\\midrule
    \lightgray{0.5} & \lightgray{300} & \lightgray{0.1}    & \lightgray{1} & 1 & \lightgray{0.1} & 5.1 \\
    \lightgray{0.5} & \lightgray{300} & \lightgray{0.1}    & \lightgray{1} & 5 & \lightgray{0.1} & 4.9 \\
    \lightgray{0.5} & \lightgray{300} & \lightgray{0.1}    & \lightgray{1} & 10 & \lightgray{0.1} & 4.9 \\\midrule
    \lightgray{0.5} & \lightgray{300} & \lightgray{0.1}    & 10  & \lightgray{1} & \lightgray{0.1} & 6.5 \\
    \lightgray{0.5} & \lightgray{300} & \lightgray{0.1}    & 50  & \lightgray{1} & \lightgray{0.1} & 8.1 \\
    \lightgray{0.5} & \lightgray{300} & \lightgray{0.1}    & 100 & \lightgray{1} & \lightgray{0.1} & 8.2 \\\midrule
    \lightgray{0.5} & \lightgray{300} & 0.001    & \lightgray{100} & \lightgray{1} & \lightgray{0.1} & 10.5 \\
    \lightgray{0.5} & \lightgray{300} & 0.0005   & \lightgray{100} & \lightgray{1} & \lightgray{0.1} & 11.3 \\
    \lightgray{0.5} & \lightgray{300} & 0.0001   & \lightgray{100} & \lightgray{1} & \lightgray{0.1} & 10.9 \\\midrule
    0.5 & 10,000 & \lightgray{0.0005} & \lightgray{100} & \lightgray{1} & \lightgray{0.1} & 9.4 \\
    0.1 & 10,000 & \lightgray{0.0005} & \lightgray{100} & \lightgray{1} & \lightgray{0.1} & 15.3 \\
    \textbf{0} & \textbf{10,000} & \textbf{0.0005} & \textbf{100} & \textbf{1} & \textbf{0.1} & \textbf{16.3} \\
    \bottomrule
  \end{tabular}
  \label{tab:appendix_sort_tuning}
\end{table}
\addtolength{\tabcolsep}{-3pt}

\subsubsection{Viterbi.}
\label{sec:appendix_tuning_viterbi}
The Viterbi approach has a number of tunable parameters.
Unfortunately, the code for this approach is prohibitively expensive to run, taking over a week of compute time to process \dataset{} train in parallel on 4 machines.
Due to this constraint, we do not tune the internal parameters of this approach.
However, \Cref{tab:appendix_viterbi_tuning} shows that tuning the tracker's per-frame score post-hoc can provide small improvements in accuracy, from 8.5 to 9.0.

\addtolength{\tabcolsep}{3pt}
\begin{table}[]
  \centering
  \caption{Results from tuning the Viterbi tracker's per-frame score threshold \dataset{} train.}
  \begin{tabular}{r@{\hskip 1em}cccccc}
    \toprule
    Tracker per-frame score & 0   & \textbf{0.1} & 0.2 & 0.3 & 0.4 & 0.5 \\ \midrule
    Track mAP       & 8.5 & \textbf{9.0} & 8.4 & 8.4 & 7.8 & 7.3
  \\\bottomrule
  \end{tabular}
  \label{tab:appendix_viterbi_tuning}
\end{table}
\addtolength{\tabcolsep}{-3pt}

\subsubsection{Tracktor++.}
\label{sec:appendix_tuning_tracktor}
Tracktor++ by default thresholds the output of a detector at 0.5. \Cref{tab:appendix_tracktor_tuning} shows the results of tuning this threshold on \dataset{} train.
Perhaps surprisingly, we find that Tracktor++ is fairly robust to this parameter, unlike SORT (as seen in \Cref{tab:appendix_sort_tuning}).
We hypothesize that this may be because of two Tracktor++ components: (1) the use of detections at time $t$ as proposal at time $t+1$ may make detectors more likely to consistently output high-confidence detections for tracks, and (2) the re-id component may allow Tracktor++ to more accurately recover tracks with no matching detections for a few frames.

\addtolength{\tabcolsep}{3pt}
\begin{table}[h!]
  \centering
  \caption{Results from tuning Tracktor's detection score threshold on \dataset{}'s train set.}
  \begin{tabular}{r@{\hskip 1em}ccccccccc}
    \toprule
  Detection score & 0.1   & 0.2   & 0.3   & \textbf{0.4}   & 0.5   & 0.6   & 0.7   & 0.8   & 0.9   \\\midrule
  Track mAP   & 35.1 & 35.5 & 35.5 & \textbf{35.7} & 35.0 & 34.7 & 34.6 & 33.0 & 29.8
  \\\bottomrule
  \end{tabular}
  \label{tab:appendix_tracktor_tuning}
\end{table}
\addtolength{\tabcolsep}{-3pt}

\subsubsection{User-initialized trackers.}
\label{sec:appendix_tuning_user_init}
As user-initialized trackers do not explicitly report when an object is \textit{absent}, we modify each method to report an object as absent when the confidence drops below a threshold.
We tune this threshold on \dataset{} train.
\Cref{tab:appendix_sot_tuning} shows that the optimal threshold varies by tracker, and tuning this parameter can lead to significant changes in accuracy (e.g., 5.2\% in the case of DIMP when using a threshold of 0.5 as opposed to the default of 0).

\addtolength{\tabcolsep}{2pt}
\begin{table}[t]
  \centering
  \caption{Results from tuning user-initialized trackers' per-frame score threshold on \dataset{} train.}
  \begin{tabular}{llllllllllll}
    \toprule
  Tracker & \multicolumn{11}{c}{Tracker per-frame score} \\
  & 0    & 0.1  & 0.2  & 0.3           & 0.4           & 0.5           & 0.6  & 0.7           & 0.8           & 0.9           & 0.99 \\ \midrule
  ATOM         & 34.3 & 34.3 & 36.2 & 36.6          & \textbf{36.7} & 34.4          & 31.9 & 26.3          & 17.8          & 9.9           & 2.1  \\
  DIMP         & 31.2 & 33.2 & 35.3 & 36.1          & 36.2          & \textbf{36.4} & 34.8 & 33.2          & 30.3          & 27.8          & 19.7 \\
  ECO          & 25.4 & 25.4 & 26.3 & \textbf{27.3} & 27.1          & 25.6          & 20.6 & 14.7          & 8.9           & 3.0           & 2.7  \\
  SiamMask     & 27.9 & 28.6 & 28.8 & 28.8          & 29.3          & 29.4          & 30.0 & 30.7          & \textbf{30.9} & 30.5          & 27.3 \\
  SiamRPN++    & 28.6 & 29.2 & 29.3 & 30.3          & 30.0          & 30.9          & 31.1 & \textbf{31.5} & 31.2          & 31.4          & 28.1 \\
  SiamRPN++-LT & 27.0 & 26.6 & 27.1 & 27.2          & 27.0          & 27.7          & 28.0 & 27.9          & 28.0          & \textbf{28.2} & 26.7 \\
  \bottomrule
  \end{tabular}
  \label{tab:appendix_sot_tuning}
\end{table}
\addtolength{\tabcolsep}{-2pt}

\begin{figure}
    \centering
    \includegraphics[width=0.85\linewidth]{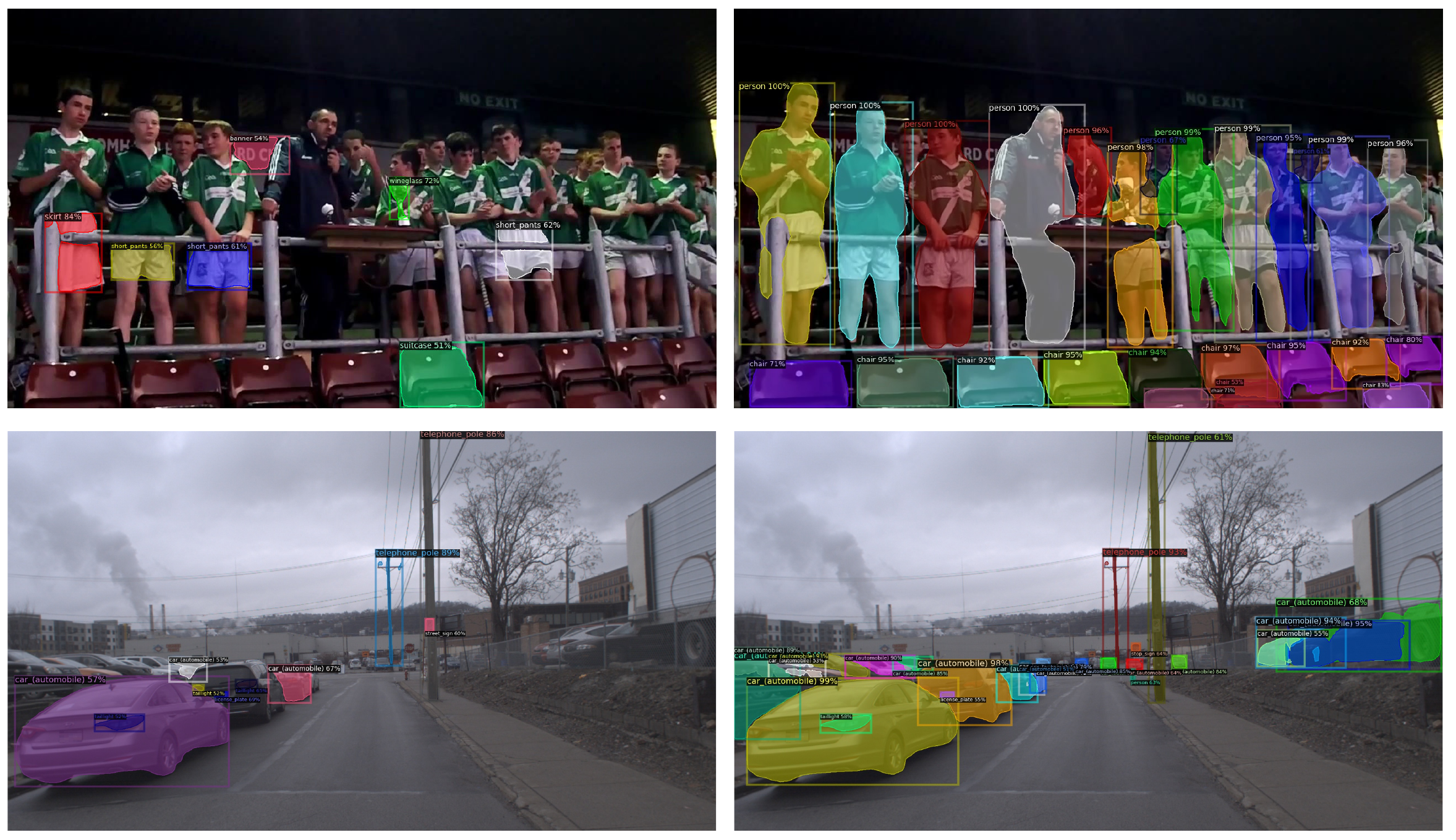}
    \caption{Qualitative comparison between a Mask R-CNN model trained on LVIS (left) and one trained on LVIS+COCO (right). Training on additional COCO data is critical for accurately detecting common categories, such as people and cars.}
    \label{fig:appendix_lviscoco}
\end{figure}

\vspace{-1cm}
\subsection{Detector details}
\label{sec:appendix_detector}

Throughout our experiments, we used a Mask R-CNN model~\cite{he2017mask} using a ResNet-101 backbone.
We re-train this model on a combination of the LVIS and COCO datasets (described below) using the
default training parameters for training on LVIS (including repeat factor sampling).
Specifically, we used the detectron2 repository~\cite{wu2019detectron2}, with the configuration file at~\url{https://github.com/facebookresearch/detectron2/blob/b6fe828a2f3b2133f24cb93c1d0d74cb59c6a15d/configs/LVIS-InstanceSegmentation/mask_rcnn_R_101_FPN_1x.yaml}.

We found that training on a combination of COCO and LVIS annotations leads to
a noticeable improvement in detection quality, which is particularly
significant for people, compared to training on LVIS alone.
To build this combination, we add COCO annotations to every image in the LVIS dataset.
To avoid duplicates, we remove COCO annotations that have IoU $>0.7$ with an LVIS annotation.
We show qualitative results of this improvement in~\Cref{fig:appendix_lviscoco}.

\dobib{}